\documentclass{article}

\usepackage{microtype}
\usepackage{graphicx}
\usepackage{subfigure}
\usepackage{booktabs} % for professional tables

\usepackage{hyperref}
\usepackage[dvipsnames]{xcolor}
\usepackage[noend]{algorithmic}
\usepackage{wrapfig}

\newcommand{\E}[2]{\mathbb{E}_{#1}{\left[#2\right]}}

\usepackage[accepted]{icml2024}

\usepackage{amsmath}
\usepackage{amssymb}
\usepackage{mathtools}
\usepackage{amsthm}

\usepackage[capitalize,noabbrev]{cleveref}

\theoremstyle{plain}
\newtheorem{theorem}{Theorem}[section]

\newtheorem{lemma}[theorem]{Lemma}
\newtheorem{corollary}[theorem]{Corollary}
\theoremstyle{definition}

\theoremstyle{remark}

\usepackage[textsize=tiny]{todonotes}

\icmltitlerunning{Learning Off-policy with Model-based Intrinsic Motivation For Active Online Exploration}

\begin{document}

\twocolumn[
\icmltitle{Learning Off-policy with Model-based Intrinsic \\ Motivation For Active Online Exploration}

% It is OKAY to include author information, even for blind
% submissions: the style file will automatically remove it for you
% unless you've provided the [accepted] option to the icml2024
% package.

% List of affiliations: The first argument should be a (short)
% identifier you will use later to specify author affiliations
% Academic affiliations should list Department, University, City, Region, Country
% Industry affiliations should list Company, City, Region, Country

% You can specify symbols, otherwise they are numbered in order.
% Ideally, you should not use this facility. Affiliations will be numbered
% in order of appearance and this is the preferred way.
% \icmlsetsymbol{equal}{*}

\begin{icmlauthorlist}
\icmlauthor{Yibo Wang}{Beihang}
\icmlauthor{Jang Zhao}{Beihang}
\end{icmlauthorlist}

\icmlaffiliation{Beihang}{School of Automation Science and Electrical Engineering, Beihang Univ., Beijing, China}

\icmlcorrespondingauthor{Jiang Zhao}{Jzhao@buaa.edu.cn}

% You may provide any keywords that you
% find helpful for describing your paper; these are used to populate
% the "keywords" metadata in the PDF but will not be shown in the document
\icmlkeywords{Reinforcement learning, Model-based Intrinsic rewards, Off-policy learning, Representation learning, Learning for control}

\vskip 0.3in
]

% this must go after the closing bracket ] following \twocolumn[ ...

% This command actually creates the footnote in the first column
% listing the affiliations and the copyright notice.
% The command takes one argument, which is text to display at the start of the footnote.
% The \icmlEqualContribution command is standard text for equal contribution.
% Remove it (just {}) if you do not need this facility.

%\printAffiliationsAndNotice{}  % leave blank if no need to mention equal contribution
\printAffiliationsAndNotice{} % otherwise use the standard text.

\begin{abstract}
Recent advancements in deep reinforcement learning (RL) have demonstrated notable progress in sample efficiency, spanning both model-based and model-free paradigms. Despite the identification and mitigation of specific bottlenecks in prior works, the agent's exploration ability remains under-emphasized in the realm of sample-efficient RL. This paper investigates how to achieve sample-efficient exploration in continuous control tasks. We introduce an RL algorithm that incorporates a predictive model and off-policy learning elements, where an online planner enhanced by a novelty-aware terminal value function is employed for sample collection. Leveraging the forward predictive error within a latent state space, we derive an intrinsic reward without incurring parameters overhead. This reward establishes a solid connection to model uncertainty, allowing the agent to effectively overcome the asymptotic performance gap. Through extensive experiments, our method shows competitive or even superior performance compared to prior works, especially the sparse reward cases.
\end{abstract}
\section{Introduction}
\label{sec:introduction}
Off-policy deep RL algorithms have always been preferred in real-world scenarios due to their ability to leverage interactive samples from diverse sources~\citep{qt-opt/18, collec_infer/22}. While these methods have attained impressive performance, substantial online interactions are inevitable, which is known as the sample efficiency problem~\citep{DPG-R/17}. As such, devising an efficient deep RL algorithm is essential for its practical application~\citep{SmithAWI/22}.

In principle, several practical ingredients have been proposed to address this issue, including increased Update-To-Data (UTD) ratio~\citep{REDQ/21}, model ensemble learning~\citep{DroQ/22}, and periodic network reset~\citep{reset-net/22}. While learning in a purely model-free manner has made strides in sample efficiency, they are inherently struggling in sparse reward cases due to the complexities of credit assignment and exploration~\citep{sac-RR/23}. In contrast, a hybrid approach, interpolating between model-based online planning and off-policy agent learning, features advantages from both sides to boost the final performance~\citep{TDMPC/22}.

Specifically, previous model-based RL methods have explored diverse ways to incorporate learned models for agent learning. Model-generated transitions are employed for data augmentation in the DynaQ framework~\citep{sutton1990integrated}. Besides, models have been used to reduce the variance of estimated policy gradient by propagating a value gradient along the environment trajectories~\citep{svg/15} or calculating analytic policy gradient throughout the model-generated rollouts~\citep{ivg/20, Dreamer_v1/20}. Model-based value expansion (MVE)~\citep{mve/18} aims to enhance value estimation at policy evaluation steps. We argue that the MVE-based method benefits the most when coupled with an online planner because of the extended lookahead horizon and improved value estimates. However, few prior works have reported further performance gain with an additional MVE-based value estimation~\citep{mbpo/19, tcrl/23}.

To address these challenges, we extend the hybrid method in two key directions. Firstly, we calculate an exponentially re-weighted average of the MVE-based temporal difference (TD) targets with different rollout horizons, striking a balance between bias and variance for accelerated credit assignment. Meanwhile, we introduce a forward predictive intrinsic reward, transforming the estimated value function into implicit exploration guidance during model-based online planning. Secondly, rather than decoupling the dynamics model and value function learning, we train them simultaneously, obtaining a planning-centric state representation.

In this work, we use this hybrid framework to build an \textbf{AC}tive online \textbf{E}xploration (ACE) planner and evaluate it on challenging high-dimensional continuous control tasks from DMControl~\citep{dmc/18}, Adroit~\citep{dapg/18} and Meta-World~\citep{metaworld/19} benchmarks, including dense and sparse reward cases. Our experimental results reveal that ACE planner consistently outperforms existing state-of-the-art algorithms,  demonstrating superior asymptotic performance across diverse environments with dense rewards. More importantly, the ACE planner excels in solving hard exploration problems and remains competitive, even in comparison to an oracle agent with access to the shaped reward function.

Compared to prior key approaches, we summarize our contributions as follows: 1. a novel hybrid model-based RL algorithm, implemented as the ACE planner, for sample-efficient RL in continuous control domains; 2. demonstration that a forward predictive measurement based on a well-shaped latent state space serves as a suitable intrinsic reward, accounting for model uncertainty; 3. introduction of an exponentially re-weighted version of the MVE target, facilitating accelerated credit assignment for off-policy agent learning; 4. theoretical and experimental validation of the benefits of a novelty-aware value function learned using the intrinsic reward for online planning.
\section{Preliminaries}
\label{sec:preliminaries}
We consider a discounted infinite horizon Markov decision process (MDP) characterized by a tuple ${(\mathcal{S, A, T, R}, \gamma, \rho_0)}$, where ${\mathcal{S} \in \mathbb{R}^n}$ and ${\mathcal{A} \in \mathbb{R}^m}$ are continuous state and action spaces, $\mathcal{T}:\mathcal{S} \times \mathcal{A} \times \mathcal{S} \to [0, \infty)$ is the state transition function, ${\mathcal{R}: \mathcal{S} \times \mathcal{A} \to \mathbb{R}}$ is a reward function, ${\gamma \in [0, 1)}$ is a discount factor, and ${\rho_0}$ the initial state distribution. The goal of an RL agent is to maximize the average discounted return given by ${j^{\rho_0}(\pi) = \mathbb{E}_{a_t \sim \pi(s_t), s_0 \sim \rho_0} [\sum^{\infty}_{t=0} \gamma^t r(s_t, a_t)]}$ under its policy ${\pi}$.

%% describe the H-step model-based online planning algos.
\subsection{Model-based Planning with Value Function}
\label{subsec:loop_description}
Integrating online planning with a terminal value function and a fully parameterized policy has demonstrated impressive performance and sample efficiency~\citep{muzero/20, efficient_zero/21}. For continuous control tasks, LOOP~\citep{loop/21} introduces a planner ${\pi^{H, \hat V }}$ which is demonstrated to be beneficial over its fixed horizon ${(\pi^{H})}$ and greedy ${(\pi^{\hat V})}$ counterparts. Given initial state ${s_0}$ and an estimated terminal value function ${\hat V}$, the desired action sequence ${\tau = (a_0, \cdots, a_{t+H-1})}$ is repeatedly optimized following:
\begin{equation}
\vspace{-1mm}
\begin{split}
    \label{eqn:MVE}
    \arg \max_{a_0, \cdots, a_{H-1}}
    \mathbb{E}_{\hat M}[\sum^{H-1}_{t=0} \gamma^{t} r(\hat s_t, a_t) + \gamma^{H} \hat V(\hat s_H)].
\end{split}
\end{equation}
Considering an approximate stochastic dynamics model ${\hat M}$ for generality, the imagined state is then sampled by ${\hat s_{t+1} \sim \hat M (s_{t+1} | s_t, a_t)}$. Typically, we can execute only the first action ${a_0}$, known as \emph{Model Predictive Control}, to get a robust closed-loop controller.

\subsection{Model-based Value Function Approximation}
\label{subsec:mve_description}
Model-based RL can build a more accurate value function estimator within the Fitted Q-iteration framework~\citep{FittedQ/05}. We refer to both state value function ${V}$ and state-action function ${Q}$ as value functions in the following for consistency. Specifically, MVE offers a cheap way to utilize the model-generated rollouts when considering a deterministic dynamics model ${\hat f}$ and gets predicted states and rewards following: ${\hat r_t = r(\hat s_t, \pi(\hat s_t))}$, ${\hat s_t = \hat f (\hat s_{t-1}, \pi(\hat s_{t-1}))}$. Given initial transition ${(s_{0}, a_{0}, r_{0}, s_{1})}$ sampled from the empirical transition distribution, the $H$-step Bellman error implemented with the TD-K trick~\citep{RL_intro/Sutton} is calculated using
\begin{equation}
\vspace{-2mm}
\begin{split}
    \label{eqn:mve_target}
    \frac{1}{H} \sum_{t=0}^{H-1} \left[ Q_{\theta}(\hat s_t, \hat a_t)-\left( \sum_{k=t}^{H-1} \gamma^{k-t} \hat r_k + \gamma^H Q_{\theta^{'}}(\hat s_H, \hat a _H) \right) \right]^2
\end{split}
\vspace{-2mm}
\end{equation}
where subscripts ${t}$ equaled to $0$ represent the sampled data and ${Q_{\theta}}$ the parameterized value function. ${\theta^{'}}$ denotes a slowly moving average of ${\theta}$, which is known as the target parameters.
\section{Curiosity Augmented Latent Space Planning}
\label{sec:latent_planning}
Planning in unknown environments for continuous control tasks requires a learned model accurately representing relevant factors. However, model bias can arise from sources such as limited state space coverage (epistemic uncertainty), transition process complexity (aleatoric uncertainty), and the choice of state features. In this section, we present the ACE planner, a model-based planning framework augmented with a novelty-aware terminal value function to mitigate epistemic uncertainty. Additionally, we optimize in a latent state space tailored to capture planning-centric representations, mitigating aleatoric uncertainty.

%% Introducing the theoretical analysis of the curiosity-augmented H-step lookahead policy
%% when compared with the greedy and no-exploration policy
\subsection{Planning with Novelty-aware Value Function}
\label{subsec:curiosity-augmented_online_planning}
We define the novelty-aware terminal value function ${V^j}$ under a modified Markov transition process ${\widetilde{\mathcal{M}}}$, where the joint reward function ${\widetilde{r} = r^e + r^i}$ evolves over environment step to represent the novelty of visited transitions. Specifically, we instantiate it as the intrinsic curiosity measured by the forward predictive error (see~\cref{sec:learning off-policy} for reward design details). With the introduction of ${V^j}$, the ACE planner ${\pi^{H, \hat V^{j}} (s_0)}$ is recovered as:
\vspace{-2mm}
\begin{equation}
\begin{split}
    \label{eqn:H-step_expl_policy}
    \arg \max_{a_0, \cdots, a_{H-1}}
    \mathbb{E}_{\hat M}[\sum^{H-1}_{t=0} \gamma^{t} r^{i}(\hat s_t, a_t) + \gamma^{H} \hat V^{j} (\hat s_H)].
\end{split}
\end{equation}
Obviously, the approximation error of ${\widetilde{\mathcal{M}}}$ and ${V^{j}}$ impacts final performance. To demonstrate the benefit of incorporating online planning with ${V^j}$, we follow previous works~\citep{ml/SinghY94, loop/21} to bound the performance of ${\pi^{H, \hat V^{j}}}$ concerning its reactive greedy counterpart ${\pi^{\hat V^{j}}(s) = \arg \textstyle \max_{a \in \mathcal{A}} \mathbb{E}_{\hat M} [\widetilde r(s, a) + \hat V^{j}(s^{'})]}$ and a local exploration policy ${\pi^{H}}$ that discards ${V^j}$. The following proofs are formulated under a stochastic dynamics model ${\hat M}$ and a state value function ${\hat V(s)}$ for generality, but the other kinds of models could be easily adopted.

%% analysis from the added reward approximated noise and the bonus q-learning bias
\begin{lemma} \label{lemma1}
Let ${\hat V^{j}}$ be approximate value function with error ${\epsilon_{v}:= \max_{s} | \hat V^{j}(s) - \widetilde{V}^{\ast}(s) |}$, where ${\widetilde{V}^{\ast}}$ is the optimal value function for MDP ${\widetilde{\mathcal{M}}}$ with the modified reward function. Let ${\pi^{\hat V^{j}(s)}}$ be the reactive greedy exploration policy. suppose curiosity bias ${\epsilon_{r} := \max_{s, a} |\widetilde{r}(s, a) - r^{e}(s, a)|}$, the performance of ${\pi^{\hat V^{j}}}$ can be bounded as:
\begin{equation}
\begin{split}
    \label{eqn:greedy_bound_main}
    J^{\pi^{\ast}} - J^{\pi^{\hat V^{j}}} \le \frac{2 \gamma (\epsilon_{v} + \alpha_r)}{1-\gamma}, \ \alpha_r = \frac{\epsilon_r}{1-\gamma}.
\end{split}
\end{equation}
\end{lemma}
%% Introduce the corollary that is based on the H-step look ahead policy's performance bound
%% and the lemma one we have drawn above.
Based on~\cref{lemma1} and the performance bound of a ${H}$-step lookahead policy derived by~\citet{loop/21} as~\cref{app:lemma2} in~\cref{appendix:proofs}, we get~\cref{corollary1} detailed below.
\begin{corollary} \label{corollary1}
Suppose ${\hat M}$ is an approximate dynamics model with Total Variation distance bounded by ${\epsilon_m}$. Let ${\hat V^j}$ be an approximate value function with error ${\epsilon_{v} := \max_{s} | \hat V^j(s) - \widetilde{V}^{\ast}(s) |}$. Let the external reward function ${r^{e}(s, a)}$ be bounded by ${[0, R^{e}_{\max}]}$ and ${\hat V^j}$ by ${[0, V^{j}_{\max}]}$. Let $\epsilon_p$ be the suboptimality incurred in the online optimization of Equation~\ref{eqn:H-step_expl_policy}. Then, the performance of the ACE planner can be bounded as:
\begin{equation}
\begin{split}
    \label{eqn:expl_bound}
    & J^{\pi^{\ast}} - J^{\pi^{H, \hat V^j}} \le \\
    & \frac{2}{1-\gamma^{H}} [C^{'}(\epsilon_m, H, \gamma) + \frac{\epsilon_p}{2} + \gamma^{H}(\epsilon_v + \alpha_r)],
\end{split}
\end{equation}
\textit{where} ${C^{'}(\epsilon_m, H, \gamma) = R^{e}_{\max} \sum^{H-1}_{t=0} \gamma^{t} t \epsilon_m + \gamma^{H} H \epsilon_m V^{j}_{\max}}$ and ${\alpha_r = \epsilon_r / (1-\gamma)}$.
\end{corollary}

\begin{proof}
We refer to~\cref{appendix:proofs} for the proof of~\cref{lemma1}, the detailed introduction of~\cref{app:lemma2}, and the extension we made to get~\cref{corollary1}.
\end{proof}
When considering the policy ${\pi^{\hat V^j}}$, its task-specific performance is highly affected by the approximation error ${\epsilon_v}$ along with the bias ${\alpha_r}$ injected by the intrinsic reward ${r^i}$. Although ${r^i}$ tends to reach zero as exploration progresses, its non-stationary nature exacerbates the value estimation accuracy. In this case, \cref{lemma1} suggests that ${\pi^{\hat V^j}}$ is ill-suited for harnessing the intrinsic reward signal, especially in the low-data regime. A corresponding perspective is confirmed by certain authors, as noted in~\citep{decouple/21}.

To solve this problem, we propose the ACE planner following insights from~\cref{corollary1}. Intuitively, the incorporation of ${\hat V^j}$ within the ${H}$-step lookahead planning framework helps diminish reliance on both the value estimation error ${\epsilon_v}$ and the injected bias ${\alpha_r}$ by a factor of ${\gamma^{H-1}}$. Despite this improvement, we must consider the degradation caused by model error ${\epsilon_m}$, where we have ${C^{'} = \mathcal{O} (R^{e}_{\max} \epsilon_m/(1- \gamma))}$ in the worst case. On the one hand, the discounted value biases mitigate the dominant error source that hinders value learning~\citep{loop/21}. On the other hand, the active exploration established by the planning process largely avoids model exploitation. As a result, our ACE planner almost addresses the asymptotic performance gap while some other model-based RL algorithms struggle. Importantly, empirical results in~\cref{sec:experiments} demonstrate its robust performance and sample efficiency, particularly in sparse reward scenarios.

\subsection{Planning with Deterministic Policy Proposal}
We instantiate our ACE planner using the Improved Cross-Entropy Method (iCEM)~\citep{icem/20}, a sample-based gradient-free planner that incorporates innovations to reduce planning budgets. Typically, planners like CEM initialize action sequences with random noise. However, we enhance its real-time efficiency by using a task-specific deterministic policy ${\pi_{\theta}}$ as the initial distribution proposal~\citep{amor/22}.

The final framework combines online planning with a model-free actor-critic RL agent. The value function is learned off-policy under the evaluation of a full parameterized policy ${\pi_{\theta}}$. We choose ${\pi_{\theta}}$ to be deterministic because we want to decouple it from exploration, thus enabling a faster convergence of the value function. Additionally, employing the policy as the distribution proposal regulates the empirical state distribution, mitigating bootstrapping errors associated with Bellman updates~\citep{loop/21}. Further details about the ACE planner are provided in~\cref{app:planner}, along with the pseudocode in~\cref{algo:ace}.
\section{Learning Off-policy with Planning-centric Representation}
\label{sec:learning off-policy}
As discussed in~\cref{sec:latent_planning}, the ACE planner operates in a latent state space for effective model learning using a planning-centric representation. This section introduces our method for representation learning and the simultaneous training of the dynamics model and RL agent, along with the derivation of the intrinsic reward.

%% introduce model components, learning objectives, intrinsic bonuses, and model uncertainties
\textbf{Agent components.}\ \ We employ a multi-step deterministic dynamics model ${d_{\theta}}$ operating in a latent space ${\mathcal{Z} \in \mathbb{R}^{m}}$, projected from physical-level states via an encoder model ${h_{\theta}}$. Given an initial latent state ${z_{t-1}}$, the model recurrently predicts the next quantities ${\hat z_{t+1} = d_{\theta}(\hat z_t, a_t, b_t), \hat r_{t+1} = r_{\theta}(\hat z_t, a_t, b_t)}$ conditioned on current predicted latent state ${\hat z_t}$, action ${a_t}$, and state belief ${b_t}$. We implement all model components as MLPs except for ${d_{\theta}}$. Actually, ${d_{\theta} := g_{\theta}(f_{\theta}(\hat z_t, a_t, b_t))}$ represents a Gate Recurrent Unit (GRU) cell~\citep{gru/14}, where ${b_{t+1} = f_{\theta}(\hat z_t, a_t, b_t)}$, followed by an MLP projector ${\hat z_{t+1} = g_{\theta}(b_{t+1})}$. To stabilize the training process, layer normalization~\citep{layernorm/16} is applied before every activation function in ${h_{\theta}}$ and ${f_{\theta}}$.

Despite suggestions from PlaNet~\citep{planet/19} advocating both deterministic and stochastic paths in latent dynamics models, the ACE planner demonstrates competence with a deterministic-only subset. Several factors contribute to this choice. Firstly, a relatively shorter predictive horizon is enough when planning with the terminal value function, alleviating model compounding errors. Besides, not only does the model complexity affect prediction accuracy, but other factors, including state feature selection, learning dynamics of the optimization process, and buffer diversity, also count. We carefully design a joint latent space and train agent components simultaneously using a multi-objective loss function, detailed in the following subsections.

\textbf{Planning-centric representation learning.} \ \ Formally, we denote a strategy that induces a state embedding ${z_t}$ to capture the intricate relationship among states, actions, and rewards as planning-centric representation learning. When the model predictive and value estimation errors are minimized conditioned on ${z_t}$, we assume this latent state is well-suited for our ACE planner.

Given this latent state ${z_t}$ and a zero-initialized state belief ${b_t}$, we define a model predictive loss over a horizon ${H}$, denoted by ${\mathcal{L}_m(\theta; \Gamma) := c_1 \mathcal{L}_d + c_2 \mathcal{L}_r}$, as follows:
\begin{equation}
\begin{split}
    \label{eqn:model_loss}
    \sum_{i=t}^{t+H}
    & c_1 \left \| \frac{q_{\theta}(d_{\theta}(\hat z_i, a_i, b_i))}
    {\left\|q_{\theta}(d_{\theta}(\hat z_i, a_i, b_i)) \right\|_2 } - 
    \frac{h_{\theta^{'}}(s_{i+1})}{\left \| h_{\theta^{'}}(s_{i+1}) \right \|_2}
    \right \|_2^{2} \\
    + & c_2 \left \| r_{\theta}(\hat z_i, a_i, b_i) - r_i
    \right \|_2^{2},
\end{split}
\end{equation}
where trajectory segments ${\Gamma = (s_t, a_t, r_t, s_t)_{t:t+H} \sim \mathcal{B}}$ are sampled from a replay buffer ${\mathcal{B}}$, ${\theta^{'}}$ represents the target parameters, and ${c_1, c_2}$ are non-negative hyper-parameters tuned for loss balance. An additional MLP predictor ${q_{\theta}}$ is employed during training for a more expressive prediction, with $\hat z_{t}$ set equal to $z_t$ for model initialization.

\cref{eqn:model_loss} employs a mean square error (MSE) between normalized predictions and target state embeddings, akin to a BYOL-style~\citep{byol/20} representation learning objective. While several self-supervised objectives are viable alternatives, we prefer $\mathcal{L}_d$ for its easy implementation and impressive performance. Authors in TCRL~\citep{tcrl/23} provide another perspective about the temporal consistency enforced by this objective. To shape a latent state space that captures planning-centric factors, the gradients from ${\mathcal{L}_m}$ are back-propagated into ${d_{\theta}}$ and ${h_{\theta}}$ across time.

\textbf{Prediction-based intrinsic reward.}\ \
The ACE planner heavily relies on an instantaneous intrinsic reward with a stable scale to reflect the novelty of the sampled trajectory segments. Leveraging the learned model, we quantify novelty via a normalized prediction error ${\ell_r (\theta, i; \Gamma_i)}$ adapted from ${\mathcal{L}_d}$ as below:
\begin{equation}
\begin{split}
\label{eqn:intrinsic_reward}
    \left \| \frac{q_{\theta}(d_{\theta}(z_i, a_i, b_i))}
    {\left\|q_{\theta}(d_{\theta}(z_i, a_i, b_i)) \right\|_2 } - 
    \frac{h_{\theta^{'}}(s_{i+1})}{\left \| h_{\theta^{'}}(s_{i+1}) \right \|_2}
    \right \|_2^{2},
\end{split}
\end{equation}
where ${\Gamma_i}$ represents transitions split from sampled trajectory segments. We reuse the model components but make one-step predictions conditioned on ${\Gamma_i}$ to reduce the reward variance. This process assigns an intrinsic reward to each transition, reflecting the complexity of its dynamics modeled at the latent level. Therefore, there exists a solid connection between ${\ell_r}$ and the model uncertainty. Finally, we introduce a reward prioritization and re-weight scheme in~\cref{app:intrinsic_reward} to account for its inherent non-stationarity.

\textbf{Off-policy learning with MVE.} \ \ To accelerate value function learning, we aim to mitigate the approximation error following the MVE approach introduced in~\cref{subsec:mve_description}. However, MVE makes strong assumptions about a nearly ideal model to hold. Besides, the distribution mismatch between visited and imagined states also degrades the performance gain. Consequently, experimental results show that it's tricky to get a successful implementation~\citep{steve/18, tcrl/23}. Nevertheless, we introduce a natural and effective integration of MVE into the off-policy learning process. Specifically, we redefine the ${H}$-step bootstrap Bellman error ${\mathcal{L}_{\lambda}(\theta; \Gamma)}$ of the value function as below:
\begin{equation}
\vspace{-2mm}
\begin{split}
\label{eqn:td_lambda_mve}
    \frac{1}{H} \sum_{i=t}^{t+H}
    \left[ Q_{\theta}(\hat z_i, a_i) - \mathcal{Q}_{\lambda} (\hat z_i, a_i) \right]^2,
\end{split}
\end{equation}
where $\mathcal{Q}_{\lambda} (\hat z_i, a_i)$ is an exponentially re-weighted average of TD targets with different rollout horizons. This revised target ${\mathcal{Q}_{\lambda} (\hat z_i, a_i)}$ can be expressed as
\begin{equation}
\vspace{-2mm}
\begin{split}
\label{eqn:td_lambda_target}
    (1 - \lambda) \sum_{k=1}^{H-1} 
    \lambda^{k-1} \mathcal{Q}^{k}(\hat z_i, a_i) + 
    \lambda^{H-1} \mathcal{Q}^{H}(\hat z_i, a_i)
\end{split}
\end{equation}
that resembles the TD${(\lambda)}$~\citep{RL_intro/Sutton} method in model-free RL literature with ${\lambda}$ balancing bias and variance. We then calculate ${\mathcal{Q}^{k}(\hat z_i, a_i)}$ within a forward horizon ${j = \min (i+k, t+H)}$ following:
\begin{equation}
\vspace{-2mm}
\begin{split}
\label{eqn:td_k_target}
    \sum_{n=i}^{j-1} \gamma^{n-i} r_n + 
    \gamma^{j-i} Q_{\theta^{'}} (h_{\theta}(s_{j}), \pi_{\theta}(h_{\theta}(s_{j})).
\end{split}
\end{equation}
Notably, we replace both the predicted rewards and terminal state in the original MVE formulation with samples from trajectory segments ${\Gamma = (s_t, a_t, r_t, s_t)_{t:t+H} \sim \mathcal{B}}$ to reduce dependency on the learned model. Meanwhile, ${z_j = h_{\theta}(s_j)}$ enforces a distribution matching that of ${z_t}$, alleviating potential distribution mismatch issue. When replacing the external reward ${r_t}$ with the joint reward ${\widetilde{r}_t = r^e_t + c_r r^i_t}$, the learned value function comes to the novelty-aware value function ${Q^{j}_{\theta}}$, with ${c_r}$ representing the constant intrinsic reward coefficient.

We employ the deterministic policy gradient schema~\citep{td3/18} to learn the policy ${\pi_{\theta}}$. It should also be noticed that the policy gradient is decoupled from the model and value function learning.

\textbf{Final algorithm} comprises two interleaved processes: active online exploration and off-policy agent learning, mutually promoting each performance. The online planner actively selects action sequences to navigate in the latent state space, prioritizing regions with high model uncertainty. After sample collection, we simultaneously train the latent dynamics model and value function via a multi-objective loss defined in~\cref{eqn:multi_objective_loss} with ${\rho}$ denoting a temporal discount factor. The loss function, combining model transition, reward prediction, and bootstrap Bellman losses, guides the joint learning process. Further details about the final algorithm and its relationship with relevant key approaches are provided in~\cref{appendix:final_algo}. The complete agent training procedure is also summarized in~\cref{algo:off-policy training}.
\vspace{-2mm}
\begin{equation}
\begin{split}
    \label{eqn:multi_objective_loss}
    \mathcal{J}(\theta; \Gamma) = \sum^{t+H}_{i=t} \rho^{i-t}
    \left( c_1 \mathcal{L}_{d, i} + c_2 \mathcal{L}_{r, i} + c_3 \mathcal{L}_{\lambda, i} \right)
\end{split}
\end{equation}
\section{Experimental Results}
\label{sec:experiments}
In this section, we first provide an intuitive understanding of the agent's exploration ability. We then evaluated our method on diverse continuous control tasks with dense and sparse reward settings to show its asymptotic performance and sample efficiency. Finally, intensive ablation studies are conducted to confirm the effectiveness of our core design choices.

\subsection{Illustration of Exploration}
\label{subsec:intuitive_expl}
\begin{figure}[t!]
    \centering
    \vspace{-0.08in}
    \includegraphics[width=0.48\textwidth]{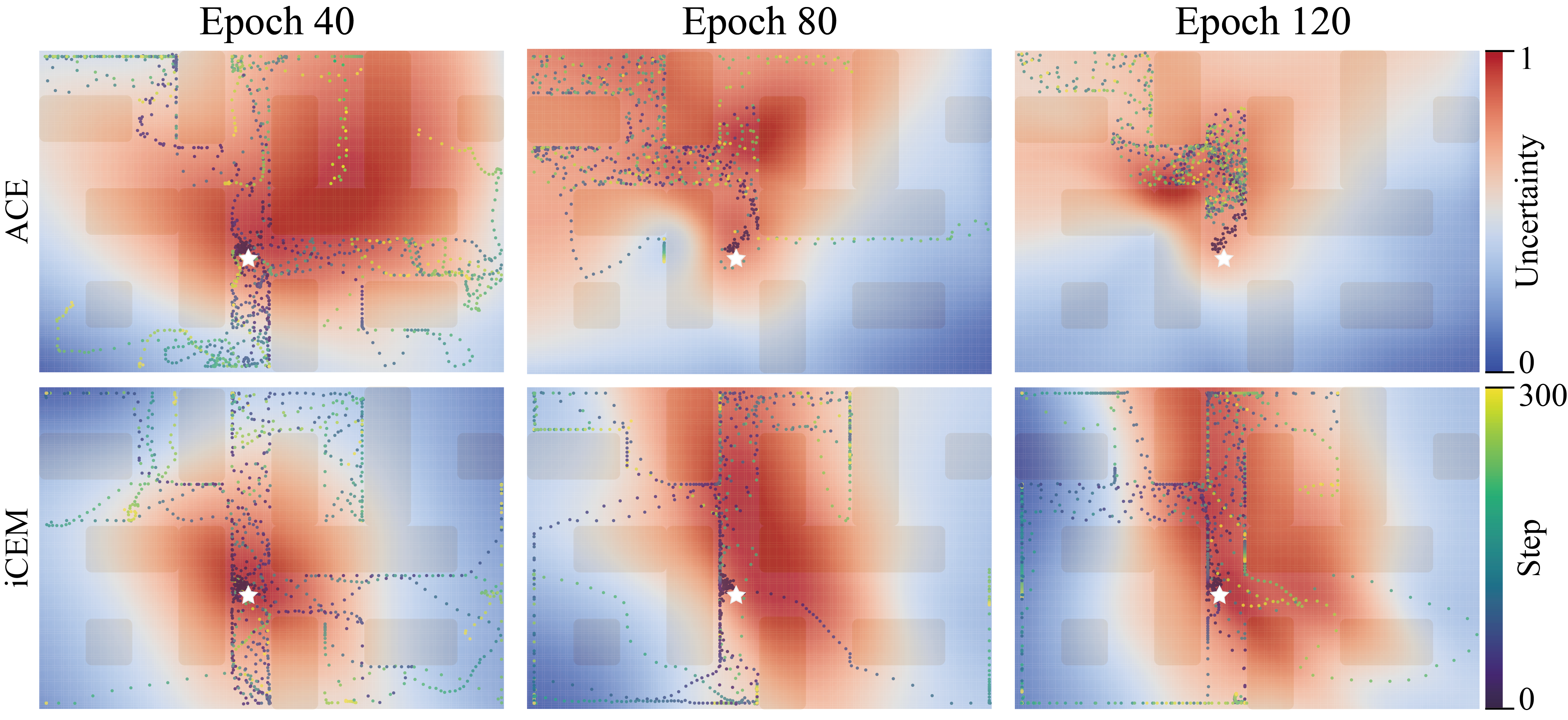}\vspace{0.10in}\\
    \includegraphics[width=0.46\textwidth]{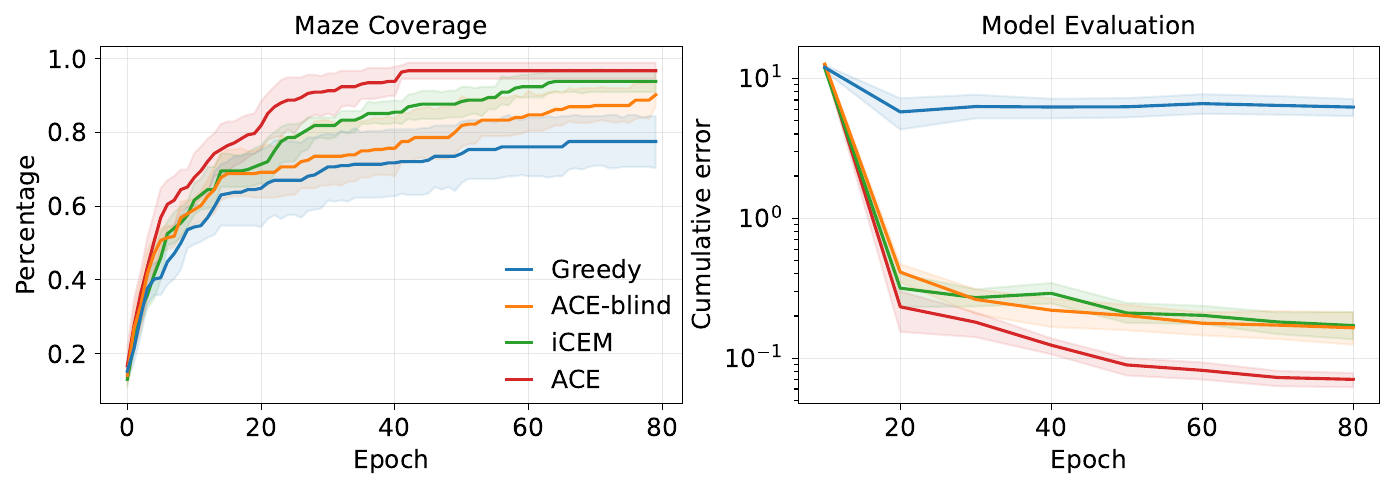}\vspace{-0.05in}
    \caption{\textbf{Exploration illustration.} \textit{(top)} Exploration traces of agents (episodic length: 300) with state novelty assessed uniformly on the maze as a grid map. The state values, calculated as the mean of predefined velocities in all directions, highlight regions with high model uncertainty (depicted in red). \textit{(Bottom)} Region coverage curve and cumulative model predictive error on the offline test set. Mean of 5 runs; shaded areas represent ${95\%}$ confidence intervals (CIs).}
    \label{fig:illustrative_traces}
\end{figure}

The ACE planner, with its stochastic yet optimized action distribution, is augmented by a novelty-aware value function to address hard exploration problems. We evaluate this enhancement in a 2D maze navigation task, depicted in~\cref{fig:illustrative_traces}. The agent, starting from a fixed position (indicated by a white star), explores the maze without external rewards. We compare the ACE planner against three exploration baselines: a vanilla iCEM planner (\textbf{iCEM}), lacking the terminal value function, and a Greedy exploration agent (\textbf{Greedy}) without lookahead search ability. Besides, we introduce \textbf{ACE-blind}, a planner without intrinsic rewards, to assess the efficiency of intrinsic reward design.

\textbf{Lookahead horizon.} Maze coverage curves in the bottom left of~\cref{fig:illustrative_traces} reveal that agents sampling actions from an optimized temporally consistent distribution exhibit superior exploration performance. The Greedy agent, despite using an intrinsic reward, explores just portions of the maze due to biased uncertainty estimation with decreased planning horizons, as suggested by~\cref{corollary1}. The intrinsic reward accelerates exploration by providing an internal signal regarding model uncertainty.

\textbf{Terminal value function.} Visualization at the top of~\cref{fig:illustrative_traces} illustrates visited trajectories and their normalized state novelty assessed by the value function. Initially, each value function assigns uniform state novelty around the start position, ensuring a diverse state distribution for off-policy agent learning. As exploration progresses, the ACE planner focuses on uncertain regions with high state values, enhancing directed exploration. In contrast, the ACE-blind agent maintains a uniform exploration pattern. Evidently, directed exploration accelerates maze coverage and value function learning, dynamically adjusting the curious region.

Finally, evaluation on a D4RL dataset of the same maze environment reveals that the ACE planner achieves the lowest model predictive error, indicating proper guidance for maze exploration provided by the novelty-aware value function. We refer to~\cref{app:d4rl_dateset} for further details on the dataset.

\subsection{Continuous Control Problems}
\textbf{Control tasks.} \ \ For experimental evaluation, we consider \textbf{25} continuous control tasks from DMControl Suite, Adroit, and Meta-World, spanning challenging locomotion, dexterous manipulation, and hard exploration tasks. We want to illustrate that the ACE planner achieves asymptotic overall performance and better sample efficiency for dense reward tasks. In the case of sparse reward conditions, we demonstrate that the ACE planner provides an effective and sample-efficient way to solve the hard exploration problems in the continuous control domain. We refer to~\cref{app:envs} for the extended description of the environment setup and evaluation metric setting.

\begin{table}[t]
\vspace{-0.1in}
\caption{\textbf{DMC15-500k benchmark results.} IQM, median, and mean performance of the ACE planner and the other five baselines on the specified DMC15-500k benchmark. ${95\%}$ bootstrap confidence intervals for results computed over 5 seeds per task. per-environment results are available in~\cref{app:dmc15_full}.}
\label{Table:IQM_DMC500k}
\begin{center}
\resizebox{0.48\textwidth}{!}{
\begin{tabular}{lccc}
\toprule
Method & IQM & Median & Mean \\
\midrule
ACE (ours) & $\mathbf{707} (698, 716)$ & $751 (741, 761)$ & $\mathbf{654} (647, 661)$ \\
TD-MPC & $668 (635, 697)$ & $706 (559, 758)$ & $625 (600, 647)$ \\
TCRL & $606 (570, 641)$ & $641 (508, 710)$ & $574 (552, 595)$ \\
REDQ & $572 (524, 617)$ & $511 (469, 541)$ & $529 (500, 556)$ \\
SR-SAC & $\mathbf{718} (704, 732)$ & $\mathbf{801} (752, 823)$ & $\mathbf{652} (641, 663)$ \\
SAC & $391 (334, 448)$ & $424 (376, 468)$ & $424 (386, 461)$ \\
\bottomrule
\end{tabular}}
\end{center}
\vspace{-0.2in}
\end{table}

\textbf{Dense reward performance.} \ \ We evaluate the ACE planner on the DMC15-500k benchmark~\citep{sac-RR/23}, a specialized DMControl version with $15$ environments and an interactive step constraint of ${5 \times 10^5}$ for a more rigorous sample efficiency comparison. A set of competitive baselines encompassing model-based and model-free RL literature are considered as follows: \textit{(1)} \textbf{TD-MPC}~\citep{TDMPC/22}, a state-of-the-art model-based RL in sample-efficiency. TD-MPC has shown consistent performance that outperforms various model-based methods. \textit{(2)} \textbf{TCRL}, a recent model-free RL algorithm demonstrating promising performance gains through a model-based temporal consistency loss. \textit{(3)} \textbf{REDQ}~\citep{REDQ/21}, a strong sample-efficient model-free RL with effective integration of regularization techniques. \textit{(4)} \textbf{SR-SAC}~\citep{sac-RR/23} makes great effort to benefit from the scaling capability of a high UTD ratio. \textit{(5)} \textbf{SAC}~\citep{sac/18} is used to highlight efficiency improvements from the learned model and planner. See~\cref {app:baseline_implementations} for baseline implementation specifics.

\cref{Table:IQM_DMC500k} presents aggregated benchmark results, focusing on metrics suggested by ~\citet{IQM/21}. The ACE planner consistently outperforms all baselines, showing its effective usage of enhanced exploration to address the asymptotic performance gap. Notably, this improvement is comparable to the gains achieved by SR-SAC with a very high UTD ratio (specifically, 128) without incurring expensive computational overhead.

\begin{figure}[t!]
    \centering
    \vspace{-0.08in}
    \includegraphics[width=0.48\textwidth, height=0.38\textwidth]{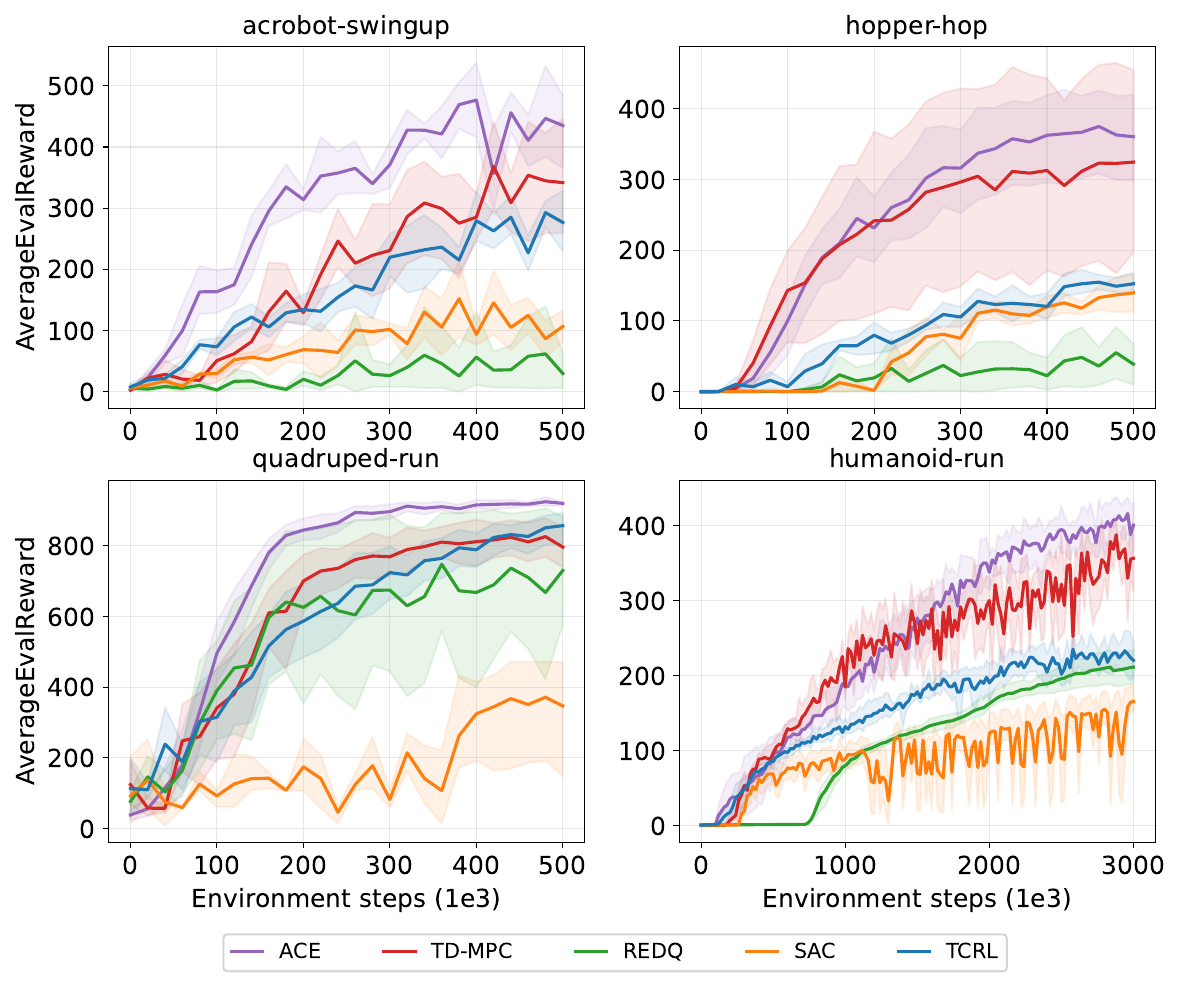}\vspace{-0.25in}
    \caption{\textbf{Learning progress in four representative tasks.} The environment steps constraint for the \textit{humanoid-run} task is relaxed to showcase the asymptotic performance achievable by each method. Mean of 5 seeds; shaded areas are $95\%$ confidence intervals.}
    \label{fig:dmc_4_learning_progress}
    \vspace{-0.20in}
\end{figure}

In \cref{fig:dmc_4_learning_progress}, we observe the learning progress across four representative tasks, providing a nuanced comparison of sample efficiency. The ACE planner demonstrates superiority in tasks with complex dynamics. Notably, in scenarios where model exploitation impedes performance, as seen in \textit{hopper-hop}, the active exploration ability helps escape local optima. When shaped rewards are competent for exploration guidance, as in REDQ and TCRL for the \textit{quadruped-run} task, the ACE planner still exhibits better sample efficiency. This is achieved by leveraging a well-learned model to accelerate credit assignment, while REDQ resorts to a high UTD ratio for a similar purpose.
Nevertheless, the proposed value expansion method outperforms REDQ significantly in challenging tasks like \textit{humanoid-run}. REDQ's reliance on a maximum entropy schema proves inefficient for exploration, exacerbated by increased bootstrap steps biasing value function learning in the early stage with a narrow transition distribution. The ACE planner avoids this issue to success, and we argue that this is the first time that an MVE-based agent achieves asymptotic performance without elaborate tricks~\citep{steve/18}. Finally, despite sharing the same BYOL-style representation learning objective, the ACE planner's active exploration schema further enhances its final performance compared to TCRL.

\textbf{Sparse reward performance.} \ \ We evaluate the ACE planner across two sparse task domains, each representing distinct exploration patterns. We denote the first as a primary goal-reaching pattern that can be solved using atomic behaviors such as periodic or symmetric motions. Instead, the other one involves multi-level or long-horizon reasoning to reach the final goal. Accordingly, two sets of tasks, including three from Adroit and six from Meta-World benchmarks, are selected, both with sparse rewards (See~\cref{app:envs} for extended environment details). Notably, for the long-horizon task set, we incorporate a hindsight experience replay (HER) buffer~\citep{her/17} to facilitate successful task completion.

In line with our prior discussion in~\cref{subsec:intuitive_expl}, we employ the same baseline methods for comparison. An evaluation metric for Adroit focuses on the normalized success counts, aligning with~\citep{rlpd/23}, to reflect task completion speed. However, the success rate metric, adhering to related works, is used for Meta-World. Besides, we include the dense reward performance of the ACE planner as an oracle for reference. Further baseline details can be found in~\cref{app:baseline_descriptions}.

\begin{figure}[t!]
    \centering
    \vspace{-0.08in}
    \includegraphics[width=0.235\textwidth]{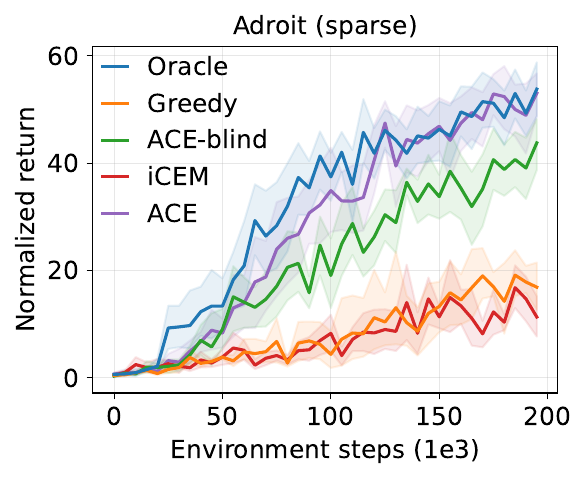} \hfill
    \includegraphics[width=0.235\textwidth]{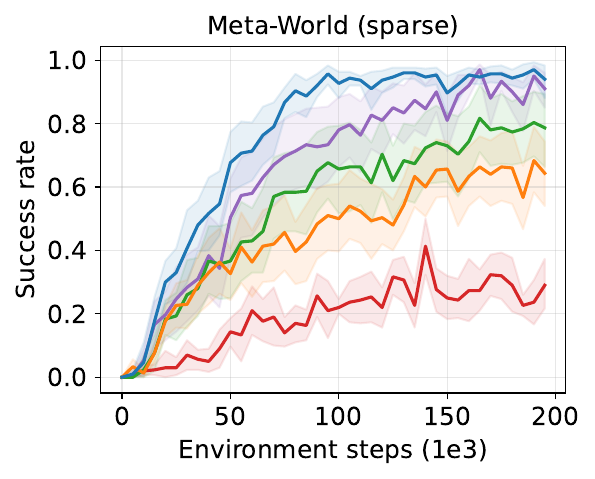}\vspace{-0.15in}
    \caption{\textbf{Aggregated sparse reward performance.} Normalized goal-reaching counts for Adroit and success rate for Meta-World as a function of environment steps, mean aggregated across individual tasks. Mean of 5 runs; shaded areas are 95\% CIs. We refer to~\cref{fig:adorit_full_results} and~\cref{fig:meta-world_full_results} for the full results.}
    \label{fig:aggregate_sparse_results}
    \vspace{-0.20in}
\end{figure}

The aggregated learning progress is summarized in~\cref{fig:aggregate_sparse_results}. Although the Adroit and Meta-World benchmarks feature distinct dynamics properties, the proposed ACE planner achieves a final performance comparable to the Oracle agent. In contrast, only the ACE-blind baseline demonstrates non-trivial results across these two benchmarks, underscoring the significance of a terminal value function for extending the lookahead search horizon. When ablating the intrinsic reward, the corresponding agent, ACE-blind, suffers certain performance degradation. By analyzing the per-task results deferred in~\cref{app:meta-world_full_results}, we observe more frequent failure cases trapped in the early exploration stage due to conservative value estimation. This argument can also be confirmed by the ablation study provided in~\cref{subsec:analysis_ablation}. Notably, the Greedy agent performs slightly better in Meta-World. This result reaffirms that an online planner is always preferred in complex dynamics domains, exactly as the Adroit compared to the Meta-World benchmark.

\subsection{Analysis and Ablation Study}\label{subsec:analysis_ablation}
\textbf{Relative importance of each component.}\ \ 
We evaluate the relative importance of core design choices by reusing the four representative tasks but replacing \textit{hopper-hop} with \textit{acrobot-swingup-sparse} to facilitate a straight comparison between dense and sparse reward cases. When ablating the MVE-based target value, improved CEM planner, and intrinsic reward, the ACE planner degenerates to \textbf{TD-MPC(S)}, a variant of TD-MPC where suffix \textbf{(S)} means the BYOL-style latent dynamics objective. Combinations of planners and target values are also explored, including \textbf{TD-MPC(S)-MVE} with an MVE-based target value and \textbf{ACE-blind-TD} featuring a one-step TD target.

\begin{figure}[t!]
    \centering
    \vspace{-0.08in}
    \includegraphics[width=0.48\textwidth, height=0.38\textwidth]{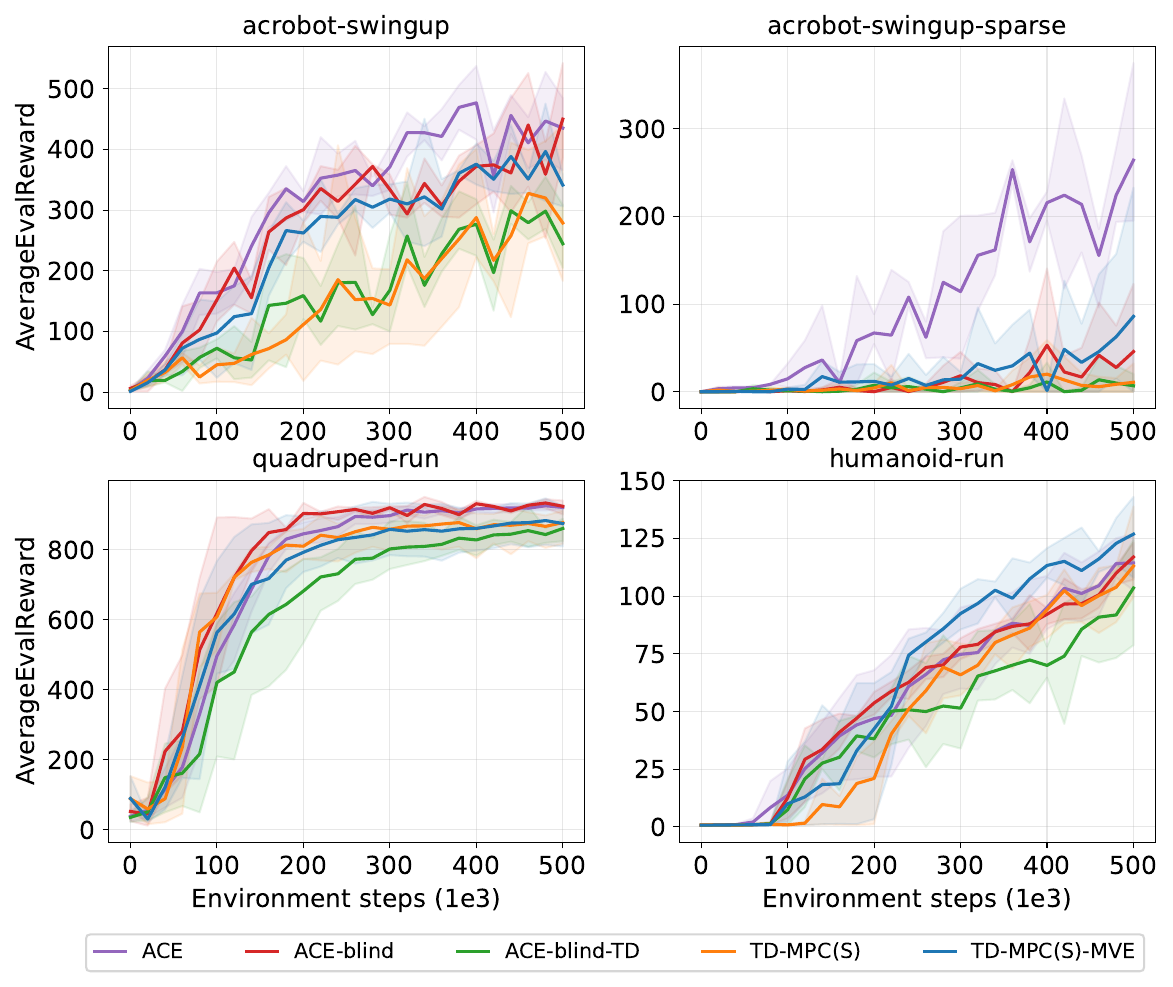}\vspace{-0.25in}
    \caption{\textbf{Relative importance of each design choice.} The MVE-based target value is the most effective factor to further improve the final performance. Mean of 5 runs; shaded areas are $95$\% CIs.}
    \label{fig:dmc_ablate_4}
    \vspace{-0.20in}
\end{figure}

\begin{figure}[t!]
    \centering
    \vspace{0.08in}
    \includegraphics[width=0.48\textwidth]{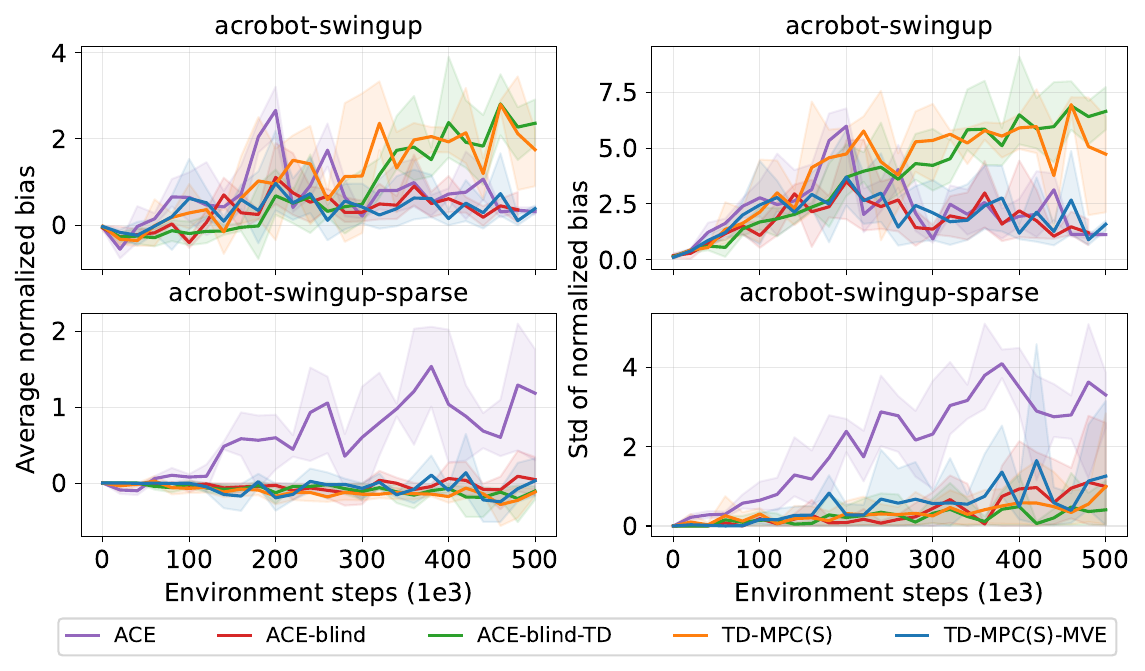}\vspace{-0.25in}
    \caption{\textbf{Average and std of the normalized estimation bias.} We showcase the results of the representative task pair to reflect the subtle difference between dense and sparse reward cases. Mean of 5 runs; shaded areas are 95\% CIs. We refer to~\cref{fig:estimate_bias_full} for full ablation results.}
    \label{fig:dmc_value_bias_main}
    \vspace{-0.20in}
\end{figure}

As depicted in~\cref{fig:dmc_ablate_4}, the intrinsic reward stands out as the most substantial improvement in \textit{acrobot-swingup-sparse} and presents no impediment when shaped rewards are available. Consequently, the intrinsic reward design can be leveraged across various continuous control tasks within the hybrid model-based RL framework. Comparison of methods employing the same value estimator but different planner types reveals that the performance gain of our approach is not derived from the iCEM planner. Instead, it is sometimes detrimental. We conjecture that this is due to the decreased sampling budgets, with $256$ in iCEM versus $512$ in CEM per iteration. Nevertheless, adopting the iCEM planner enhances the practicality of our method in real-world scenarios where real-time property is critical. Notably, replacing the one-step TD target with an MVE-based target value emerges as the most effective factor in dense reward tasks and, when combined with the intrinsic reward, in sparse reward tasks.

\textbf{Does MVE improve the value estimation?}\ \
To probe the quality of value estimation, we employ the Monte Carlo method to replicate the average and standard deviation (std) of the normalized value estimation bias~\citep{REDQ/21}. The average bias indicates whether the target value is overestimated or underestimated, while std measures the uniformity of the estimation bias.

We focus on a task pair consisting of \textit{acrobot-swingup} and \textit{acrobot-swingup-sparse} where the intrinsic reward is shown to make a positive but distinct impact on value function learning. In~\cref{fig:dmc_value_bias_main}, we observe that the MVE-based target value is crucial for mitigating the overestimation and maintaining a lower normalized std that avoids ambiguous action preference~\citep{doubleQ/16}. It is noteworthy that certain overestimation in the early training phase does not impede final performance but can be seen as an implicit factor promoting state space exploration. As a result, agents without intrinsic reward in the \textit{acrobot-swingup-sparse} task get trapped in local optima because of the conservative value estimation.

Moreover, we note that the sample efficiency in the sparse reward task is inferior to the dense one. Due to the spare learning signal and a long-term credit assignment process, the value estimation bias decreases slower with a corresponding higher normalized std, exacerbating the stability of value function learning. This suggests a future research direction about how to mitigate value overestimation with sparse reward signals in a sample-efficient manner. We defer the full results featuring the same trends in~\cref{app:extended_ablation_results}.

\textbf{Is HER necessary for long-horizon tasks?}

\begin{wrapfigure}{r}{0.20\textwidth}
    \centering
    \vspace{-5mm}
    \includegraphics[width=0.20\textwidth,trim={2mm 0 1mm 0}, clip]{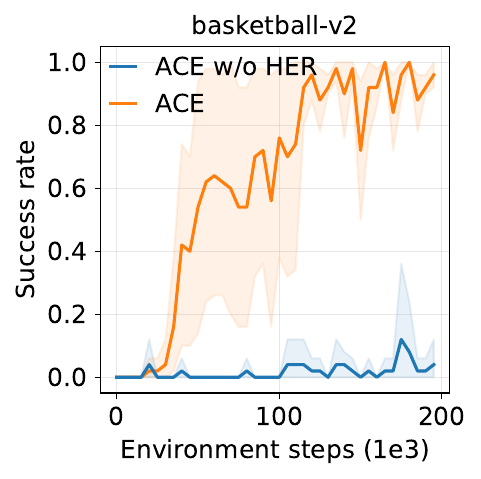} \\
    \vspace{-4mm}
    \caption{A success rate comparison between the ACE planner with or without the HER buffer.}
    \label{fig:her_ablation}
    \vspace{-3mm}
\end{wrapfigure}
Primary experimental results on \textit{basketball-v2} task, as illustrated in~\cref{fig:her_ablation}, reveal that the ACE planner itself is struggling to solve the long-horizon goal-reaching problems. We visualize the failure mode and find that the agent most frequently gets stuck in the object manipulation stage, which can be seen as a local optimal condition. The HER buffer relabels intermediate states as sub-goals to solve this issue. However, challenges arise in tasks with multiple intermediate subgoals, where the HER buffer may be less effective in the absence of a hierarchical structure. Consequently, incorporating a sub-goal generator into the ACE planner emerges as a promising avenue for sample-efficient exploration in more intricate tasks.
\section{Future Work and Limitations}
\label{sec:furtrue_works}
This paper focuses on sample-efficient exploration for continuous control tasks and relies on the interaction of multiple critical components to work. However, opportunities for further enhancements and limitations are acknowledged. 

Our current online planner optimizes action sequences with a limited horizon, which is deemed to be short-sighted~\citep{iql_tdmpc/23}. Consequently, the HER buffer should be incorporated to ensure the success of long-horizon decision-making tasks. A promising way for improvement involves modifying the planner to operate hierarchically, where temporal abstractions of the environment can be leveraged for sub-goal reaching~\cite{director/22, iql_tdmpc/23}. 

While the predictive intrinsic reward demonstrates simplicity and a solid connection to the uncertainty of the latent dynamics model, challenges arise in the presence of uncontrollable factors, such as the noisy-TV problem~\citep{curiosity_rl/19}, in stochastic environments. Structuring the latent state space with a set of object-centric representations excludes task-irrelevant noise and makes the directed exploration interpretable~\citep{cswm/20, slot_attention/20}. Besides, certain representation learning techniques~\citep{denoised_mdp/22, SEAR/23} that yield disentangled state embeddings could extend the proposed method to noisy environments. Furthermore, the integration of innovation architectures, like transformers~\citep{TDM/23} and Structured State Space Sequence (S4) Model~\citep{s4model/23}, into planning and off-policy agent learning remains an open research problem for further investigation.
\section{Conclusion}
\label{conclusion}
In this study, we present the ACE planner, a model-based RL algorithm that seamlessly integrates online planning and off-policy agent learning for sample-efficient exploration. Drawing on theoretical insights gained from the performance analysis of an $H$-step lookahead policy coupled with predictive intrinsic rewards, we enhance the ACE planner with a novelty-aware terminal value function learned using the exponential reweighted MVE-based value estimator. By employing the active exploration schema, our approach demonstrates remarkable sample efficiency in mastering challenging continuous control tasks, particularly those with no external reward signal.
\newpage
\section*{Impact Statements}
This paper presents a model-based RL agent for sample efficient exploration, which promotes the application of deep RL algorithms in real-world scenarios. The intrinsic reward, measured by model predictive error, serves as a learning signal without introducing bias to the final performance, as facilitated by the proposed MVE-based value estimator. Looking ahead, the ACE planner could be extended to act in real environments and simultaneously learn to become skilled in the tasks it encounters in a short time frame.

Nevertheless, deploying the agent on real robots poses challenges, primarily related to the absence of a secure data collection framework. Besides, the reward mismatch issue may lead to undesired agent behaviors. Future advances in uncertainty quantification, multi-level hierarchical decision-making, and safety-aware policy learning will improve our ability to apply the learning-based control method to risky real-world control tasks.

\bibliography{references}
\bibliographystyle{icml2024}
%%%%%%%%%%%%%%%%%%%%%%%%%%%%%%%%%%%%%%%%%%%%%%%%%%%%%%%%%%%%%%%%%%%%%%%%%%%%%%%
%%%%%%%%%%%%%%%%%%%%%%%%%%%%%%%%%%%%%%%%%%%%%%%%%%%%%%%%%%%%%%%%%%%%%%%%%%%%%%%
% APPENDIX
%%%%%%%%%%%%%%%%%%%%%%%%%%%%%%%%%%%%%%%%%%%%%%%%%%%%%%%%%%%%%%%%%%%%%%%%%%%%%%%
%%%%%%%%%%%%%%%%%%%%%%%%%%%%%%%%%%%%%%%%%%%%%%%%%%%%%%%%%%%%%%%%%%%%%%%%%%%%%%%
\newpage
\appendix
\onecolumn
\section{Proofs} \label{appendix:proofs}
This section provides proofs for the performance bounds introduced in the main paper.
We begin by adopting an upper bound established by \citet{ml/SinghY94} in a specific case where the MDP is modified using the prediction-based intrinsic reward, presented as a lemma.

\begin{lemma} \label{app:lemma1}
Let ${\hat V^{j}}$ be an approximate value function with error ${\epsilon_{v}:= \max_{s} | \hat V^{j}(s) - \widetilde{V}^{\ast}(s) |}$, where ${\widetilde{V}^{\ast}}$ is the optimal value function for the MDP ${\widetilde{\mathcal{M}}}$ with modified reward function. Let ${\pi^{\hat V^{j}}(s) = \arg \max_a \mathbb{E}_{s^{'} \sim M(\cdot | s, a)}[\widetilde{r}(s, a) + \gamma \hat{V}^{j}(s^{'})]}$ be the reactive greedy exploration policy. suppose curiosity bias ${\epsilon_{r} := \max_{s, a} |\widetilde{r}(s, a) - r^{e}(s, a)|}$, the performance of ${\pi^{\hat V^{j}}}$ can be bounded as:
\begin{equation}
\vspace{-2mm}
\begin{split}
\label{eqn:greedy_bound}
    L_{\pi^{\hat V^{j}}} := J^{\pi^{\ast}} - J^{\pi^{\hat V^{j}}} \le \frac{2 \gamma (\epsilon_{v} + \alpha_r)}{1-\gamma}, \ \alpha_r = \frac{\epsilon_r}{1-\gamma}.
\end{split}
\end{equation}
\end{lemma}

\begin{proof} \label{app:proof_lemma1}
We define ${x}$ as a state with the maximum performance gap, i.e., ${\forall s \in \mathcal{S}, L_{\pi^{\hat V^{j}}}(x) \ge L_{\pi^{\hat V^{j}}} (s)}$. Let ${u = \pi^{\ast}(x)}$ be the optimal action for state ${x}$ under the original MDP, and ${a = \pi^{\hat V^{j}}(x)}$ be the one-step greedy action specified by ${\hat V^{j}}$. When evaluating action ${u}$ under the approximate value function ${\hat V^{j}}$, it must be inferior to the one-step greedy action ${a}$ as:
\begin{equation}
\begin{split}
    \label{eqn:state_evalution}
    \widetilde{r}(x, u) + \gamma \sum_{x^{'}}[p(x^{'} | x, u) \hat V^{j}(x^{'})] \le 
    \widetilde{r}(x, a) + \sum_{x^{'}}[p(x^{'} | x, a) \hat V^{j}(x^{'})].
\end{split}
\end{equation}
Given the joint reward function ${\widetilde{r} = r^{e} + r^{i}}$, where the intrinsic reward ${r^{i}}$ is constrained to ${[0, \epsilon_r]}$, we establish an upper bound on the gap between the optimal value functions ${\widetilde{V}^{\ast}}$ and ${V^{\ast}}$ as:
\begin{equation}
\begin{split}
    \label{eqn:optimality_gap}
    \left|\widetilde{V}^{\ast}(s) - V^{\ast}(s) \right| = & \left| \mathbb{E}_{\widetilde{\tau}^{\ast} \sim p_{\widetilde{\tau}^{\ast}}} \left[ \sum \gamma^{t}\widetilde{r}(s_t, a_t) \right] - \mathbb{E}_{\tau^{\ast} \sim p_{\tau^{\ast}}} \left[ \sum \gamma^{t}r^{e}(s_t, a_t) \right] \right| \\
    \le & \left| \mathbb{E}_{\widetilde{\tau}^{\ast} \sim p_{\widetilde{\tau}^{\ast}}} \left[ \sum \gamma^{t}(\widetilde{r}(s_t, a_t) - r^{e}(s_t, a_t)) \right] \right| \\
    \le & \ \frac{\epsilon_r}{1-\gamma} \\
    = & \ \alpha_r.
\end{split}
\end{equation}
When combining~\cref{eqn:optimality_gap} with value approximation error bound ${\epsilon_v}$, we reach to a conclusion that for all ${x^{'} \in \mathcal{S}}$, ${V^{\ast}(x^{'}) - \epsilon_{v} - \alpha_r \le \hat {V}^{j}(x^{'}) \le V^{\ast}(x^{'}) + \epsilon_{v} + \alpha_r}$. We substitute this conclusion into~\cref{eqn:state_evalution} and then get:
\begin{equation}
\begin{split}
    \label{eqn:optimal_state_evaluation}
    \widetilde{r}(x, u) + \gamma \sum_{x^{'}}[p(x^{'} | x, u) V^{\ast}(x^{'}) - \epsilon_{v} - \alpha_r] \le 
    \widetilde{r}(x, a) + \gamma \sum_{x^{'}}[p(x^{'} | x, a) V^{\ast}(x^{'}) + \epsilon_{v} + \alpha_r].
\end{split}
\end{equation}
Combining~\cref{eqn:optimal_state_evaluation} with the non-negativity of the intrinsic reward, we deduce that
\begin{equation}
\begin{split}
    \label{eqn:optimal_state_evaluation_trans}
    r^{e}(x, u) - \widetilde{r}(x, a) \le & \ 
    2 \gamma (\epsilon_{v} + \alpha_r) - r^{i}(x, u) + \gamma \sum_{x^{'}}[p(x^{'} | x, a) V^{\ast}(x^{'}) - p(x^{'} | x, u) V^{\ast}(x^{'})] \\
    \le & \ 2 \gamma (\epsilon_{v} + \alpha_{r}) + \gamma \sum_{x^{'}}[p(x^{'} | x, a) V^{\ast}(x^{'}) - p(x^{'} | x, u) V^{\ast}(x^{'})].
\end{split}
\end{equation}
Formally, the performance bound we consider here is
\begin{equation}
\begin{split}
    L_{\pi^{\hat V^{j}}}(x) = & J^{\pi^{\ast}} - J^{\pi^{\hat V^j}} \\
    = & r^{e}(x, u) - \widetilde{r}(x, a) + \gamma \sum_{x^{'}}[p(x^{'} | x, u) V^{\ast}(x^{'}) - p(x^{'} | x, a) \hat V^j(x^{'})],
\end{split}
\end{equation}
Substituting~\cref{eqn:optimal_state_evaluation_trans} yields
\begin{equation}
\begin{split}
    L_{\pi^{\hat V^j}}(x) \le & \ 2 \gamma (\epsilon_{v} + \alpha_r) + \gamma \sum_{x^{'}}[p(x^{'} | x, a) V^{\ast}(x^{'}) - p(x^{'} | x, a) \hat V^j(x^{'})] \\
    = & \ 2 \gamma (\epsilon_{v} + \alpha_r) + \gamma \sum_{x^{'}} p(x^{'} | x, a) L_{\pi^{\hat V^j}}(x^{'}).
\end{split}
\end{equation}
Given our assumption ${\forall s \in \mathcal{S}, L_{\pi^{\hat V^j}}(x) \ge L_{\pi^{\hat V^j}} (s)}$, we conclude that
\begin{equation}
\begin{split}
    L_{\pi^{\hat V^j}}(x) \le 2 \gamma (\epsilon_{v} + \alpha_r) + \gamma \sum_{x^{'}} p(x^{'} | x, a) L_{\pi^{\hat V^j}}(x).
\end{split}
\end{equation}
Finally, we get 
\begin{equation}
\begin{split}
    L_{\pi^{\hat V^j}}(x) \le \frac{2 \gamma (\epsilon_{v} + \alpha_r)}{1-\gamma}, \ \alpha_r = \frac{\epsilon_r}{1-\gamma}.
\end{split}
\end{equation}
\end{proof}

To analyze the performance of the proposed ACE planner, we would like to briefly present the performance bound for the ${H}$-step lookahead policy introduced in~\citet{loop/21} as stated in~\cref{app:lemma2}.
\begin{lemma} \label{app:lemma2}
(\citet{loop/21}) Suppose ${\hat M}$ is an approximate dynamics model with Total Variation distance bounded by ${\epsilon_m}$. Let ${\hat V}$ be an approximate value function such that ${\mathrm{max}_s  \left| V^{\ast}(s) - \hat{V}(s) \right| \le \epsilon_v}$. Let the reward function ${r(s, a)}$ be bounded by ${[0, R_{\max}]}$ and ${\hat V}$ be bounded by ${[0, V_{\max}]}$. Let $\epsilon_p$ be the suboptimality incurred in H-step lookahead optimization (\ref{eqn:H-step_expl_policy}). Then the performance of the $H$-step lookahead policy ${\pi^{H, \hat V}}$ can be bounded as:
\begin{equation}
\begin{split}
    \label{eqn:bound_of_loop}
    J^{\pi^{\ast}} - J^{\pi^{H, \hat V}} \le \frac{2}{1-\gamma^{H}} [C(\epsilon_m, H, \gamma) +
    \frac{\epsilon_p}{2} + \gamma^{H} \epsilon_v],
\end{split}
\end{equation}
\vspace{-2mm}
\textit{where}
\vspace{-2mm}
\begin{equation*}
C(\epsilon_m, H, \gamma) = R_{\max} \sum^{H-1}_{t=0} \gamma^{t} t \epsilon_m + \gamma^{H} H \epsilon_m V_{\max}
\end{equation*}
\end{lemma}

Based on~\cref{app:lemma1} and~\cref{app:lemma2}, we establish the performance bound for the ACE planner, as presented in~\cref{app:corollry1}. This corollary showcases the effectiveness of the ACE planner in harnessing intrinsic rewards for sample-efficient exploration under the Fitted-Q iteration framework~\citep{FittedQ/05}. Formally, the corollary is formulated as:
\begin{corollary} \label{app:corollry1}
Suppose ${\hat M}$ is an approximate dynamics model with Total Variation distance bounded by ${\epsilon_m}$. Let ${\hat V^j}$ be an approximate value function with error ${\epsilon_{v} := \max_{s} | \hat V^j(s) - \widetilde{V}^{\ast}(s) |}$. Let the external reward function ${r^{e}(s, a)}$ be bounded by ${[0, R^{e}_{\max}]}$ and ${\hat V^j}$ by ${[0, V^{j}_{\max}]}$. Let $\epsilon_p$ be the suboptimality incurred in the online optimization of Equation~\ref{eqn:bound_of_ace}. Then, the performance of the ACE planner can be bounded as:
\begin{equation}
\begin{split}
\label{eqn:bound_of_ace}
    J^{\pi^{\ast}} - J^{\pi^{H, \hat V^j}} \le
    \frac{2}{1-\gamma^{H}} [C^{'}(\epsilon_m, H, \gamma) + \frac{\epsilon_p}{2} + \gamma^{H}(\epsilon_v + \alpha_r)],
\end{split}
\end{equation}
\vspace{-2mm}
\textit{where}
\vspace{-2mm}
\begin{equation*}
    C^{'}(\epsilon_m, H, \gamma) = R^{e}_{\max} \sum^{H-1}_{t=0} \gamma^{t} t \epsilon_m + \gamma^{H} H \epsilon_m V^{j}_{\max} \ \ \mathrm{and} \ \ \alpha_r = \epsilon_r / (1-\gamma)
\end{equation*}
\end{corollary}

\begin{proof}
Consider an approximate value function with error bound ${\epsilon_{v} := \max_{s} | \hat V^j(s) - \widetilde{V}^{\ast}(s) |}$ and an approximate dynamics model with Total Variation distance ${D_{\mathrm{TV}} (M(\cdot | s, a), \hat M(\cdot | s, a)) \le \epsilon_m}$ as the model deviation metric, similar to the measurement in~\cite{mbpo/19}. We intend to analyze the impact of approximation errors in both the value function and dynamics model on the ACE planner's performance gap.

We adopt the notations from~\citep{loop/21} for a consistent formulation. Accordingly, let ${\mathcal{M}}$ be an MDP defined by ${(\mathcal{S}, \mathcal{A}, M, r, s_0)}$, where ${M}$ is the ground truth dynamics, ${\mathcal{S}}$ the state space, ${\mathcal{A}}$ the action space, $r$ the reward function, and $s_0$ the initial state. When ${M}$ is substituted with an approximate dynamics model ${\hat M}$ in ${\mathcal{M}}$, we have the MDP ${\hat{\mathcal{M}}}$ defined by ${(\mathcal{S, A}, \hat M, r, s_0)}$. Correspondingly, let ${\mathcal{H}}$ be an $H$-step finite horizon MDP characterized by ${(\mathcal{S}, \mathcal{A}, M, r_{\mathrm{mix}}, s_0)}$, and ${\hat{\mathcal{H}}}$ be an $H$-step finite MDP given with ${(\mathcal{S}, \mathcal{A}, \hat{M}, r_{\mathrm{mix}}, s_0)}$, in which
\begin{equation}
    r_{\mathrm{mix}}(s_t, a_t) = \begin{cases}
        r(s, a) &\mathrm{if} \ \ t<H \\
        \hat{V}^{j}(s_H) &\mathrm{if} \ \ t=H
    \end{cases} \\
\end{equation}
Assuming ${\pi^{H, \hat V^{j}}}$ is the ACE planner optimized online for the $H$-step lookahead objective, formulated by~\cref{eqn:H-step_expl_policy}, in ${\hat{\mathcal{H}}}$ while acting for $H$ steps in ${\mathcal{M}}$. We follow the same setting as~\citep{loop/21} in which the ACE planner operates in a shooting manner to ease the proof. We also want to highlight that the intrinsic reward has no effect on the online optimization procedure except for the terminal value function because we fit a reward function using just the ground truth labels to exclude factors that will bias the overall performance.

Denoting the optimal policy in a specific MDP ${\mathcal{K}}$ as ${\pi^{\ast}_{\mathcal{K}}}$, let ${\hat \tau}$, ${\tau}$, and ${\tau^{\ast}}$ represent $H$-step trajectories sampled by running ${\pi^{\ast}_{\hat{\mathcal{H}}}}$, ${\pi^{\ast}_{\mathcal{H}}}$, and ${\pi^{\ast}_{\mathcal{M}}}$ in ${\mathcal{M}}$ respectively.  Correspondingly, let ${p_{\hat \tau}, p_{\tau}}$, and ${p_{\tau^{\ast}}}$ be the trajectory distributions. The considered performance gap is defined by the following:
\begin{align}
    &J^{\pi^*}-J^{\pi^{H,\hat{V}^j}} = V^*(s_0) - V^{\pi^{H,\hat{V}^j}} (s_0)\\
    &=\E{\tau^*\sim p_{\tau^*}}{\sum\gamma^t r(s_t,a_t) +\gamma^H V^*(s_H)} - \E{\hat{\tau}\sim p_{\hat{\tau}}}{\sum\gamma^tr(s_t,a_t) +\gamma^H V^{\pi^{H,\hat{V}^j}}(s_H)}\\
    &=\E{\tau^*\sim p_{\tau^*}}{\sum\gamma^tr(s_t,a_t) +\gamma^H V^*(s_H)} -\E{\hat{\tau}\sim p_{\hat{\tau}}}{\sum\gamma^tr(s_t,a_t) +\gamma^H V^*(s_H)}\\
    &+\E{\hat{\tau}\sim p_{\hat{\tau}}}{\sum\gamma^tr(s_t,a_t) +\gamma^H V^*(s_H)}- \E{\hat{\tau}\sim p_{\hat{\tau}}}{\sum\gamma^tr(s_t,a_t) +\gamma^H V^{\pi^{H,\hat{V}^j}}(s_H)}\\
    \label{eq:ineq0}
    &=\E{\tau^*\sim p_{\tau^*}}{\sum\gamma^tr(s_t,a_t) +\gamma^H V^*(s_H)} -\E{\hat{\tau}\sim p_{\hat{\tau}}}{\sum\gamma^tr(s_t,a_t)+\gamma^H V^*(s_H)}\\
    &+ \gamma^H\E{\hat{\tau}\sim p_{\hat{\tau}}}{V^*(s_H)-V^{\pi^{H,\hat{V}^j}}(s_H)}
\end{align}
Because we have demonstrated that ${| V^{\ast}(s) - \hat{V}^{j}(s) | \le \epsilon_v + \alpha_r, \forall s \in \mathcal{S}}$, we can bound the following expressions:
\begin{align}
\label{eq:ineq1}
    &\E{\tau^*\sim p_{\tau^*}}{\sum\gamma^tr(s_t,a_t) +\gamma^H V^*(s_H)} \le \E{\tau^*\sim p_{\tau^*}}{\sum\gamma^tr(s_t,a_t) +\gamma^H \hat{V}^{j}(s_H)}+\gamma^H(\epsilon_v + \alpha_r)\\
\label{eq:ineq2}
    &\E{\hat{\tau}\sim p_{\hat{\tau}}}{\sum\gamma^tr(s_t,a_t) +\gamma^H V^*(s_H)} \ge \E{\hat{\tau}\sim p_{\hat{\tau}}}{\sum\gamma^tr(s_t,a_t) +\gamma^H \hat{V}^{j}(s_H)}-\gamma^H (\epsilon_v + \alpha_r)
\end{align}
Subtracting these two equations reaches an inequality:
\begin{align}
\label{eq:ineq3}
    &\E{\tau^*\sim p_{\tau^*}}{\sum\gamma^tr(s_t,a_t) +\gamma^H V^*(s_H)} -\E{\hat{\tau}\sim p_{\hat{\tau}}}{\sum\gamma^tr(s_t,a_t) +\gamma^H V^*(s_H)} \\
    \le &\E{\tau^*\sim p_{\tau^*}}{\sum\gamma^tr(s_t,a_t) +\gamma^H \hat{V}^{j}(s_H)} \nonumber -\E{\hat{\tau}\sim p_{\hat{\tau}}}{\sum\gamma^tr(s_t,a_t) +\gamma^H \hat{V}^{j}(s_H)} + 2\gamma^H(\epsilon_v + \alpha_r)
\end{align}
When substituting~\cref{eq:ineq3} into~\cref{eq:ineq0}, we can get the transformed performance bound as follows:
\begin{align}
    &J^{\pi^*}-J^{\pi^{H,\hat{V}^{j}}}=V^*(s_0) - V^{\pi^{H,\hat{V}^{j}}} (s_0)\nonumber\\
    &\le \E{\tau^*\sim p_{\tau^*}}{\sum\gamma^tr(s_t,a_t) +\gamma^H \hat{V}^{j}(s_H)}  -\E{\hat{\tau}\sim p_{\hat{\tau}}}{\sum\gamma^tr(s_t,a_t) +\gamma^H \hat{V}^{j}(s_H)} \\
    &+ 2\gamma^H(\epsilon_v + \alpha_r) \nonumber + \gamma^H\E{\hat{\tau}\sim p_{\hat{\tau}}}{V^*(s_H)-V^{\pi^{H,\hat{V}^{j}}}(s_H)}\\
    &= \E{\tau^*\sim p_{\tau^*}}{\sum\gamma^tr(s_t,a_t) +\gamma^H \hat{V}^{j}(s_H)}-\E{\tau \sim p_{\tau}}{\sum\gamma^tr(s_t,a_t) +\gamma^H \hat{V}^{j}(s_H)}\\
    &+\E{\tau\sim p_{\tau}}{\sum\gamma^tr(s_t,a_t) +\gamma^H \hat{V}^{j}(s_H)} \nonumber -\E{\hat{\tau}\sim p_{\hat{\tau}}}{\sum\gamma^tr(s_t,a_t) +\gamma^H \hat{V}^{j}(s_H)} \\
    &+ 2\gamma^H(\epsilon_v + \alpha_r) + \gamma^H\E{\hat{\tau}\sim p_{\hat{\tau}}}{V^*(s_H)-V^{\pi^{H,\hat{V}^{j}}}(s_H)}\\
    \label{eq:ineq8}
    &\le \E{\tau\sim p_{\tau}}{\sum\gamma^tr(s_t,a_t) +\gamma^H \hat{V}^{j}(s_H)}  -\E{\hat{\tau}\sim p_{\hat{\tau}}}{\sum\gamma^tr(s_t,a_t) +\gamma^H \hat{V}^{j}(s_H)} \\
    &+ 2\gamma^H(\epsilon_v + \alpha_r) + \gamma^H\E{\hat{\tau}\sim p_{\hat{\tau}}}{V^*(s_H)-V^{\pi^{H,\hat{V}^{j}}}(s_H)}
\end{align}
The last inequality is derived from the fact that optimal action sequences, evaluated in the ground truth MDP ${\mathcal{M}}$, exhibit inferior performance when applied in the $H$-step MDP ${\mathcal{H}}$. This implies $\E{\tau^*\sim p_{\tau^*}}{\sum\gamma^tr(s_t,a_t) +\gamma^H \hat{V}^{j}(s_H)}\le\E{\tau \sim p_{\tau}}{\sum\gamma^tr(s_t,a_t) +\gamma^H \hat{V}^{j}(s_H)}$. Actually, the expression in~\cref{eq:ineq8} represents the performance gap between optimal policies ${\pi^{\ast}_{\mathcal{H}}}$ of MDP ${\mathcal{H}}$ and ${\pi^{\ast}_{\hat{\mathcal{H}}}}$ of MDP ${\hat{\mathcal{H}}}$, both evaluated in MDP ${\mathcal{H}}$.

To streamline the proof, we adopt the performance gap conclusion from~\citep{loop/21}:
\begin{equation*}
   \E{\tau\sim p_{\tau}}{\sum\gamma^t r(s_t,a_t) +\gamma^H \hat{V}(s_H)}  -\E{\hat{\tau}\sim p_{\hat{\tau}}}{\sum\gamma^t r(s_t,a_t) +\gamma^H \hat{V}(s_H)} \le 2C(\epsilon_m,H,\gamma)+\epsilon_p.
\end{equation*}
Here, the discrepancy measurement is given by ${C = R_{\mathrm{max}} \sum^{H-1}_{t=0} \gamma^t t \epsilon_m + \gamma^H H \epsilon_m V_{\mathrm{max}}}$, primarily reflecting the value approximation error in the last additive term. Since we employ an $H$-step MDP ${\mathcal{H}}$ sharing the same dynamics model but substituting only the terminal reward with the novelty-aware value function, we conclude that the considered performance gap can be expressed as:
\begin{equation}
\label{eqn:primative_bound}
   \E{\tau\sim p_{\tau}}{\sum\gamma^t r(s_t,a_t) +\gamma^H \hat{V}^{j}(s_H)}  -\E{\hat{\tau}\sim p_{\hat{\tau}}}{\sum\gamma^t r(s_t,a_t) +\gamma^H \hat{V}^{j}(s_H)} \le 2C^{'}(\epsilon_m,H,\gamma)+\epsilon_p,
\end{equation}
in which ${C^{'} = R^{e}_{\mathrm{max}} \sum^{H-1}_{t=0} \gamma^t t \epsilon_m + \gamma^H H \epsilon_m V^{j}_{\mathrm{max}}}$.

Finally, we substitute~\cref{eqn:primative_bound} in~\cref{eq:ineq8} and use the recursive property of 
\begin{equation}
\begin{split}
    J^{\pi^*}-J^{\pi^{H,\hat{V}^{j}}} &= V^*(s_0) - V^{\pi^{H,\hat{V}^{j}}} (s_0) \\
    &\le 2C^{'}(\epsilon_m,H,\gamma) + \epsilon_p + \gamma^H\E{\hat{\tau}\sim p_{\hat{\tau}}}{V^*(s_H)-V^{\pi^{H,\hat{V}^{j}}}(s_H)}
\end{split}
\end{equation}
to get that 
\begin{equation}
\begin{split}
    J^{\pi^*}-J^{\pi^{H,\hat{V}^{j}}} \le \frac{2}{1-\gamma^H}
    \left[ C^{'}(\epsilon_m, H, \gamma) + \frac{\epsilon_p}{2} + \gamma^H (\epsilon_v + \alpha_r) \right]
\end{split}
\end{equation}
\end{proof}

In summary, we formulate~\cref{app:corollry1} as a specific instance of the performance bound for the $H$-step lookahead policy. This instance involves substituting the terminal reward function of the $H$-step MDP with a novelty-aware value function. It provides us with a theoretical tool to analyze the impacts introduced by the intrinsic reward under the $H$-step online optimization schema. Consequently, the ACE planner, as a model-based RL agent, demonstrates efficacy in achieving sample-efficient exploration and asymptotic performance across diverse continuous control tasks.

\section{Planner} \label{app:planner}
We instantiate our ACE planner using iCEM~\citep{icem/20}, an enhanced CEM planner chosen for its real-time efficiency during both training and inference procedures. As discussed in~\cref{sec:latent_planning}, we conduct trajectory optimization in a latent state space ${\mathcal{Z}}$ to take advantage of the planning-centric representations.

During planning initiation, parameters ${(\mu^0, \sigma^0)_{t:t+H}, \mu^0, \sigma^0 \in \mathbb{R}^m, \mathcal{A} \in \mathbb{R}^m}$ for the action sequences ${\tau}$ are initialized as the mean and standard deviation of a colored noise distribution with exponent ${\beta}$. To adopt the iCEM planner for efficient latent space planning, we combine noise-sampled and policy-proposed action sequences for latent dynamics model rollouts in each update iteration. These policy-proposed action sequences help narrow down the search space because of their task-specific properties.

After sample evaluation, the top-$E$ sequences ${(\tau^{\ast}_{0:E})}$, ranked by their total returns ${\phi_{\tau}^{\ast}}$, are selected as an elite-set to update ${\mu^k}$ and ${\sigma^k}$ via an exponential average at the ${k}$-th iteration following:
\begin{equation}
\begin{split}
\label{eqn:gaussian_fit}
    \mu^k = \frac{\sum^E_{i=1} \Omega_i \tau_i^{\ast}} 
    {\sum^E_{i=1} \Omega_i},
    \sigma^k = \sqrt{\frac{\sum^E_{i=1} \Omega_i (\tau_i^{\ast} - \mu^k)^2}
    {\sum^E_{i=1} \Omega_i}},
\end{split}
\end{equation}
where the total return of a model rollout ${\phi_{\tau}}$ is estimated by $\mathbb{E}_{\hat M}[\sum^{H-1}_{t=0} \gamma^t r^{e}(\hat z_t, a_t) + \gamma^H \hat V^j (\hat z_H)]$, and ${\Omega_i = e^{\phi^{\ast}_{\tau_i} / \eta}}$ is the reweighted factor controlled by a temperature coefficient ${\eta}$.

This optimization procedure is repeated for ${K}$ iterations with the reuse of the elite-set ${\tau^{\ast}}$, decay of the sampled population size ${S_k}$, and other improvements introduced in iCEM. We refer to~\cref{algo:ace} for the detailed operations. After the final iteration, only the first action of the best elite ${\tau^{\ast}_0}$ is executed in the environment. Utilizing the instantiated model components in~\cref{sec:learning off-policy}, we summarize the complete planning procedure in~\cref{algo:ace}.

\begin{figure}[!t]
\centering
\resizebox{0.78\linewidth}{!}{%
\begin{minipage}{0.78\linewidth}
\begin{algorithm}[H]
\caption{iCEM planner for active online exploration}
\label{algo:ace}
\begin{algorithmic}[1]
\REQUIRE latent dynamics model ${d_{\theta}}$, reward model ${r_{\theta}}$; \\
~~~~~~~~~~number of iteration ${K}$, reduction factor of samples ${\psi}$; \\
~~~~~~~~~~planning horizon ${H}$, number of samples ${S}$, elite fraction ${E}$; \\
~~~~~~~~~~proposal policy ${\pi_{\theta}}$, novelty-aware value function ${Q^{j}_{\theta}}$, state encoder ${h_{\theta}}$; \\
~~~~~~~~~~state ${s_0}$, colored-noise exponent ${\beta}$, noise mean and strength ${(\mu_{init}, \sigma_{init})}$. \\
\FOR{each time-step $t=0 \ldots T-1$}
    \IF{t == 0}
    \STATE ${\mu_t \gets \mu_{init} \in \mathbb{R}^{m, H}}$
    \ELSE
    \STATE ${\mu_{t} \leftarrow}$ shifted ${\mu_{t-1}}$ (repeat for last time-step)
    \ENDIF
    \STATE ${\sigma_t \gets \sigma_{init}}$
    \FOR{each iteration ${k=0 \ldots K-1}$}
    \STATE ${S_k \gets \max (S \cdot \psi^{-k}, 2 \cdot E)}$
    \hfill \textit{\color{CadetBlue}// Apply decay of population size with a lower bound ${E}$} \\
    {\textit{\color{CadetBlue}// Sample action from the noise distribution}}
    \STATE samples ${\gets S_k}$ samples from ${\mathrm{clip} (\mu_t + \mathcal{C}^{\beta}(m,H) \odot \sigma^2_t)}$
    \IF{k == 0}
    \STATE add fraction of shifted elite-set ${\tau^{\ast}}$ from last time-step to samples
    \ELSE
    \STATE add fraction of shifted elite-set ${\tau^{\ast}}$ from last iteration to samples
    \ENDIF
    \STATE ${(s_t, a_t, \ldots, s_{t+H}) \gets \mathrm{rollout\_with\_policy}(d_{\theta}, \pi_{\theta}, H)}$
    \STATE samples ${\gets (a_0, a_1, \ldots, a_H)}$
    \hfill \textit{\color{CadetBlue}// Add policy proposed action into the samples}
    \IF{k == K-1}
    \STATE add ${\mu_t}$ to samples
    \hfill \textit{\color{CadetBlue}// Add mean of the samples at the last iteration}
    \ENDIF
    \STATE ${ \phi_{\tau} \gets \mathrm{evaluate\_action\_samples} (s_t, \tau, d_{\theta}, H, r_{\theta}, Q^{j}_{\theta}) }$
    \STATE ${\tau^{\ast} \gets \mathrm{best} \  E \ \mathrm{samples\ according\ to\ ranked}\ \phi_{\tau}}$
    \STATE ${\mu_t, \sigma_t \gets}$ fit Gaussian distribution to ${\tau^{\ast}}$ with momentum according to Equation~\ref{eqn:gaussian_fit}
    \ENDFOR
    \STATE executes the first action of the best elite ${\tau^{\ast}_{0}}$
\ENDFOR
\end{algorithmic}
\end{algorithm}
\end{minipage}%
}
\end{figure}

\section{Reward Prioritization and Reweighing}
\label{app:intrinsic_reward}
To account for the inherent non-stationarity of the intrinsic reward, we employ a normalization and prioritization schema akin to BYOL-Explore~\citep{byol_explore/22}. Additionally, we incorporate Min-Max normalization with an exponent ${\xi}$ to maintain a consistent intrinsic reward scale throughout the training process. Although this transformation makes the intrinsic reward non-zero, empirical results indicate that there is no negative impact on the final performance.

Specifically, the raw intrinsic reward, denoted as ${\ell_r(\theta, i)}$, is normalized by its Exponential Moving Average (EMA) estimate of the standard deviation ${\sigma_r}$. To prioritize transitions with the highest uncertainties, we follow BYOL-Explore by clipping the normalized reward ${\ell_r(\theta, i) / \sigma_r}$ using a reward threshold ${\mu_r / \sigma_r}$, where ${\mu_r}$ is the EMA estimate of the mean reward. Finally, we scale the clipped reward ${r_i = \mathrm{max}((\ell_r(\theta, i)- \mu_r) / \sigma_r, 0)}$ using the reweighted maximum value as follows:
\begin{equation}
\begin{split}
\label{eqn:reward_nrom_scale}
    r_{\max} = \max_i [(\ell_r(\theta, i)- \mu_r) / \sigma_r, 0], \ \
    r_i = r_i ^ {-\xi} / r_{\max} ^ {-\xi}.
\end{split}
\end{equation}

\section{Further Algorithm Details}
\label{appendix:final_algo}
\subsection{Extended Description of ACE planner}
We elaborate on the final algorithm, comparing it to relevant counterparts, with a focus on sample efficiency, including sparse-reward scenarios. Recent advancements in regularization techniques, such as dropout, feature normalization, and network reset, have empowered model-free RL agents in high-dimensional control domains~\citep{avtd/23}. However, it's notable that most model-free RL methods focus on dense reward cases, leaving sparse reward tasks relatively less explored. Empirical results from the main paper and some prior works~\citep{TDMPC/22, sac-RR/23} have shown that the exploration ability is a primary bottleneck for further improvement.

\begin{table}[t]
\caption{\textbf{Comparison with key approaches.} We summarize essential components attributing to the sample efficiency of leading model-based and model-free methods, emphasizing crucial design choices for detailed comparison. \textit{Efficiency booster} indicates the source of sample efficiency gain, \textit{regularization term} refers to essential learning techniques preventing issues like overestimation~\citep{td3/18} and primacy~\cite{reset-net/22} biases, \textit{exploration} denotes behavior policies for interactive sample collection, and \textit{compute} offers a rough estimate of the relative computational cost during training and inference. The term repre. abbreviates representation for brevity.}
\label{Table:relevant_works}
\vspace{0.05in}
\centering
\resizebox{0.8\textwidth}{!}{%
\begin{tabular}{lcccc}
\toprule
\textbf{Method} & Efficiency booster & Exploration & Regularization term & Compute \\ \midrule
REDQ & High UTD ratio (20) & Maximum entropy & Model ensemble & Moderate \\
SR-SAC & High UTD ratio (128) & Maximum entropy & Periodic reset & High \\ 
TCRL & Temporal consistent repre. & Gaussian noise & Layer normalization & Low \\ \midrule
TD-MPC & Planning-centric repre. & CEM w/ policy & Discount factor & Moderate \\
Dreamer & Analytical policy gradient & Gaussian noise & TD-lambda & Moderate \\
LOOP & Lookahead ability & CEM w/ policy & Model Ensemble & Moderate \\
RPG & Planning-centric repre. & Multimodal distribution & Discount factor & Low \\ \midrule
\textbf{ACE} (ours) & Planning-centric repre. w/ MVE & iCEM w/ policy & TD-lambda & Moderate \\ \bottomrule
\end{tabular}
}
\end{table}

Most recently, \citet{sparse_redq/23} extends the REDQ algorithm to sparse-reward tasks, emphasizing the importance of combining REDQ with an HER buffer and proper regularization for sample-efficient exploration. These results also show a promising way to extend its model-based counterparts.

In contrast, model-based RL methods employ internal dynamics models for sample efficiency. Thus, they are more flexible in incorporating components designed for efficient learning, such as the DynaQ framework, model-based value expansion, and improved policy gradient calculation. Nevertheless, model bias is always inevitable, which makes it hard to achieve asymptotic performance. In these cases, model ensemble~\citep{mbpo/19, planet/19} and some other uncertainty quantification methods can be leveraged to mitigate the model error or even further serve as guidance for environment exploration~\citep{max/19, avtive_mani/22}.

As a model-based RL agent, the ACE planner operates within a Tandem setting~\citep{tandem/21}, employing an off-policy agent that learns exclusively from data collected by the online planner. Unlike methods relying on a high UTD ratio for accelerated value function learning, the ACE planner utilizes a learned model for the MVE-based value target calculation, enhancing computational efficiency. To overcome the exploration bottleneck, the ACE planner incorporates a novelty-aware terminal value function, implicitly guiding the environment exploration. Nevertheless, several prior works combine the model-based RL with model-free RL elements. The most similar method, TD-MPC, plans with a terminal value function but without access to the intrinsic reward signal. This makes it struggle under some sparse reward tasks.

Notably,~\citet{rpg/23} proposes a principled framework that models the continuous RL policy as a multi-modal distribution. The implemented RL method, denoted as reparmeterized policy gradient (RPG), leverages the learned model and an intrinsic reward based on random network distillation (RND)~\citep{rnd/19} to achieve sample-efficient exploration. In principle, our proposed MVE-based value estimator can be incorporated into the RPG framework to enhance its performance further. Besides, the active exploration ability derived from the online planner tends to be critical to account for model uncertainty. More future work is required to make a comprehensive comparison between these two different exploration schemas.

Finally,~\cref{Table:relevant_works} provides an element-wise comparison of the most relevant approaches. The complete off-policy training process is also shown in~\cref{algo:off-policy training}.

\begin{figure}[ht]
\centering
\resizebox{0.75\linewidth}{!}{%
\begin{minipage}{0.75\linewidth}
\begin{algorithm}[H]
\caption{Learning off-policy with the ACE planner}
\label{algo:off-policy training}
\begin{algorithmic}[1]
\REQUIRE initialized network parameters ${\theta, \theta^{'}}$, rollout horizon ${H}$; \\
~~~~~~~~~~learning rate ${\xi}$, rollout discount ${\rho}$, constant loss coefficients ${c_1, c_2, c_3}$; \\
~~~~~~~~~~prioritized replay buffer ${\mathcal{B}}$ with exponents ${(\alpha, \beta)}$, intrinsic reward coefficient ${c_r}$.
\WHILE{continue training}
    \FOR{step ${t = 0 \ldots T}$}
    \STATE ${a_t \sim \pi^{H, Q^{j}_{\theta}}(s_t)}$ \hfill \textit{\color{CadetBlue} // Sample with the online planner}
    \STATE ${(s_{t+1}, r_t) \sim \mathcal{T}(\cdot | s_t, a_t)}$
    \STATE ${\mathcal{B} \leftarrow \mathcal{B} \cup (s_t, a_t, r_t, s_{t+1})}$
    \ENDFOR
    \textit{\color{CadetBlue} // Update the dynamics model and off-policy agent simultaneously}
    \FOR{update iteration}
    \STATE ${(s_t, a_t, r_t, s_{t+1})_{t:t+H+1} \sim \mathrm{PER}(\mathcal{B}}; \alpha, \beta)$ \hfill \textit{\color{CadetBlue} // Sampling from PER buffer}
    \STATE ${z_{t:t+H}, z^{'}_{t:t+H+1} = h_{\theta}(s_{t:t+H}), h_{\theta^{'}}(s_{t:t+H+1})}$ \hfill \textit{\color{CadetBlue} // Extract latent states}
    \STATE ${\ell_r(\theta, i) = \ell_{r, \theta}(z_t, z^{'}_{t+1})_{t:t+H}}$ \hfill \textit{\color{CadetBlue} // Calculate raw intrinsic reward by~\cref{eqn:intrinsic_reward}}
    \STATE ${r^{inc}_{t:t+H} \gets \mathrm{norm}(\ell_r(\theta, i), \mathrm{count} = 1)}$ \hfill \textit{\color{CadetBlue} // Apply normalization by~\cref{eqn:reward_nrom_scale}}
    \STATE ${J, J_{\pi} = 0, 0 \quad \hat z_t = z_t} $
        \FOR{rollout horizon ${i = t \ldots t+H}$}
        \STATE ${\hat r_i = r_{\theta}(\hat{z}_i, a_i)}$
        \STATE ${\widetilde r_i = r_i + c_r \cdot r^{inc}_i}$  \hfill \textit{\color{CadetBlue} // Prepare the joint reward for value learning}
        \STATE ${\hat q_i = Q^{j}_{\theta}(\hat{z}_i, a_i)}$
        \STATE ${\hat{z}_{i+1} = d_{\theta}(\hat{z}_i, a_i)}$ \hfill \textit{\color{CadetBlue} // Rollout the latent dynamics model}
        \STATE ${J \leftarrow J + \rho^{i-t}} \mathcal{L}(z^{'}_{i+1}, \widetilde{r}_i, \hat q_i)$ \hfill \textit{\color{CadetBlue} // Calculate weighted total loss by~\cref{eqn:multi_objective_loss}}
        \STATE ${J_{\pi} \gets J_{\pi} - Q^{j}_{\theta}(\hat{z}_i, \pi_{\theta}(\hat{z}_i))}$ \hfill \textit{\color{CadetBlue} // Calculate deterministic policy gradient}
        \ENDFOR
    \STATE ${\theta \leftarrow \theta - \frac{1}{H} \xi \bigtriangledown_{\theta} J (J_{\pi})}$ \hfill \textit{\color{CadetBlue} // Update the online networks}
    \STATE ${\theta^{'} \leftarrow (1-\tau) \theta^{'} + \tau \theta}$ \hfill \textit{\color{CadetBlue} //  Update the target networks}
    \ENDFOR
\ENDWHILE
\end{algorithmic}
\end{algorithm}
\end{minipage}%
}
\end{figure}

\subsection{Baselines}\label{app:baseline_descriptions}
In this section, we describe the implementation details of the baseline methods for a fair comparison.

\textbf{TD-MPC}. We obtain the reported results for state-based DMControl tasks from~\citet{TDMPC/22}. For additional tasks within the DMC15-500k benchmark, we adapt the official implementation with default hyper-parameters, adjusting only the action repeat to match the configuration tuned by the ACE planner for consistency.

\textbf{REDQ}. We leverage the publicly available implementation but change the benchmark environments from Mujoco to DMControl. We set the UTD ratio as 20 across all tasks, as further increments often result in performance saturation or degradation.

\textbf{iCEM}. A vanilla iCEM~\citep{icem/20} planner with a learned latent dynamics model but discards the terminal value function to clarify its implicit guidance effect on exploration. The iCEM agent shares the same off-policy learning setting, maintaining consistent hyper-parameters with the proposed ACE planner.

\textbf{Greedy}. We replace the online planner of the ACE agent with its fully parameterized internal policy to create the Greedy exploration agent. Notably, it behaves as a reactive policy due to its lack of lookahead search ability. We compare this greedy agent to the ACE planner to show the efficiency derived from the active exploration paradigm.

\textbf{ACE-blind}. A variant of the ACE planner trained without access to the intrinsic reward, allowing us to evaluate the effectiveness of the intrinsic rewards design. We maintain the learning recipe consistent across all three exploration baselines, ensuring they benefit from the planning-centric state representation.

\textbf{Other baselines}. Benchmark results reported in related works for \textbf{TCRL}~\citep{tcrl/23}, \textbf{SAC}~\citep{sac/18}, and \textbf{SAC-RR}~\citep{sac-RR/23} are included. The unified benchmark and open-source efforts allow us to focus on the algorithmic design aspects.

\subsection{Implementation Details}\label{app:baseline_implementations}
In~\cref{sec:learning off-policy}, we outlined the model components and the final algorithm.
In this section, we provide the implementation details of our method. Because the ACE planner shares algorithm schema with TD-MPC, we reuse most hyper-parameters and model structures from TD-MPC's implementation. Firstly, we maintain the policy ${\pi_{\theta}}$ and reward head ${r_{\theta}}$ as two-layer MLPs. Then, the encoder model ${h_{\theta}}$ is modified to be a one-layer MLP with layer normalization. This adaptation proves effective in handling state features with diverse scales. We also apply layer normalization to the value function ${Q_{\theta}}$ and the single-layer GRU cell ${f_{\theta}}$ to ensure training stability. Consistent with prior work~\citep{byol/20}, we introduce the dynamics projector ${g_{\theta}}$ and predictor ${q_{\theta}}$ as one-layer MLPs with batch normalization, preventing representation collapse.

\cref{Table:ace hyper-parameters} provides an overview of all relevant hyper-parameters for both planning and off-policy learning. Additionally, three task-specific hyper-parameters are considered: action repeat, colored-noise exponent, and the balance coefficient for TD targets. Refer to~\cref{Table:per_task_settings} for these specific values.

\begin{table}[h!]
\caption{\textbf{ACE planner hyper-parameters.} We here list all the relevant hyper-parameters for the off-policy learning and online planning sub-processes.}
\label{Table:ace hyper-parameters}
\centering
\resizebox{0.6\textwidth}{!}{%
\begin{tabular}{ll} \toprule
    \textbf{Hyper-parameters} & \textbf{Value}  \\ \midrule
    \textbf{Off-policy learning} &  \\
    Discount factor ${(\gamma)}$ & 0.99 \\
    Seed episodes & 5 \\
    Rollout horizon & 6 (DMControl, Meta-World) \\
     & 7 (Adroit) \\
    Rollout discount & 0.2 (pen) \\
     & 0.5 (otherwise) \\
    Replay buffer size & Unlimited \\
    Sampling schema & PER $(\alpha=0.6, \beta=0.4)$ \\
    MLP hidden size & 512 \\
    MLP activation & ELU \\
    GRU hidden size & 128 \\
    Encoder hidden size & 256 \\
    Latent dimension & 100 (Humanoid, Adroit) \\
     & 50  (otherwise) \\
    Learning rate & 1e-3 \\
    Optimizer & AdamW ${\beta_1 = 0.9, \beta_2 = 0.999}$ \\
    Similarity loss coefficient & 1.0 \\
    Reward loss coefficient & 0.5 \\
    Value loss coefficient & 0.1 \\
    Intrinsic reward coefficient & ${0.25}$ (dense) \\
     & ${0.50}$ (sparse) \\
    Batch size & 512 \\
    Target networks Momentum & 0.99 \\
    Policy delay & 2 iterations \\
    \midrule
    \textbf{Online planning} & \\
    Planning horizon ${(H)}$ & 6 (DMControl, Meta-World) \\
     & 7 (Adroit) \\
    Horizon schedule & ${2 \to 6}$ (5 episodes) \\
    Iterations & 6 \\
    Initial mean and variance ${(\mu, \sigma)}$ & (0, 0.5) \\
    Variance lower bound & ${0.5 \to 0.05}$ (5 episodes) \\
    Population size ($S$) & 700 (Adroit), 256 (otherwise) \\
    Elite fraction ($E$) & 32 \\
    Reduction factor ${\psi}$ & 1.25 \\
    Fraction reused elites & 0.25 \\
    Policy fraction & 0.5 \\
    Mean momentum coefficient & 0.1 \\ 
    Score temperature & 0.5 \\ \bottomrule
\end{tabular}%
}
\end{table}

\begin{figure}[ht!]
    \centering
    \vspace{0.05in}
    \includegraphics[width=0.184\textwidth]{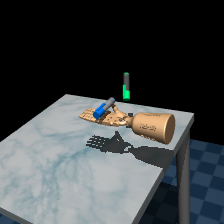}
    \includegraphics[width=0.184\textwidth]{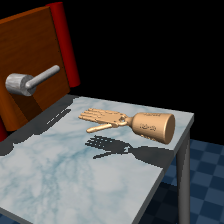}
    \includegraphics[width=0.184\textwidth]{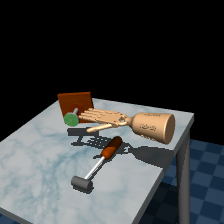}
    \includegraphics[width=0.184\textwidth]{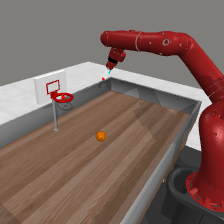} 
    \includegraphics[width=0.184\textwidth]{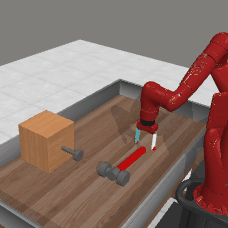} \vspace{0.05in} \\
    \includegraphics[width=0.375\textwidth, height=0.184\textwidth]{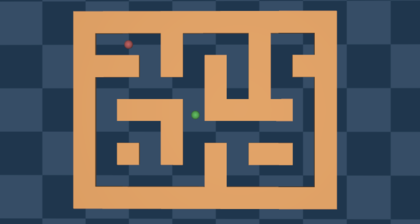}
    \includegraphics[width=0.184\textwidth]{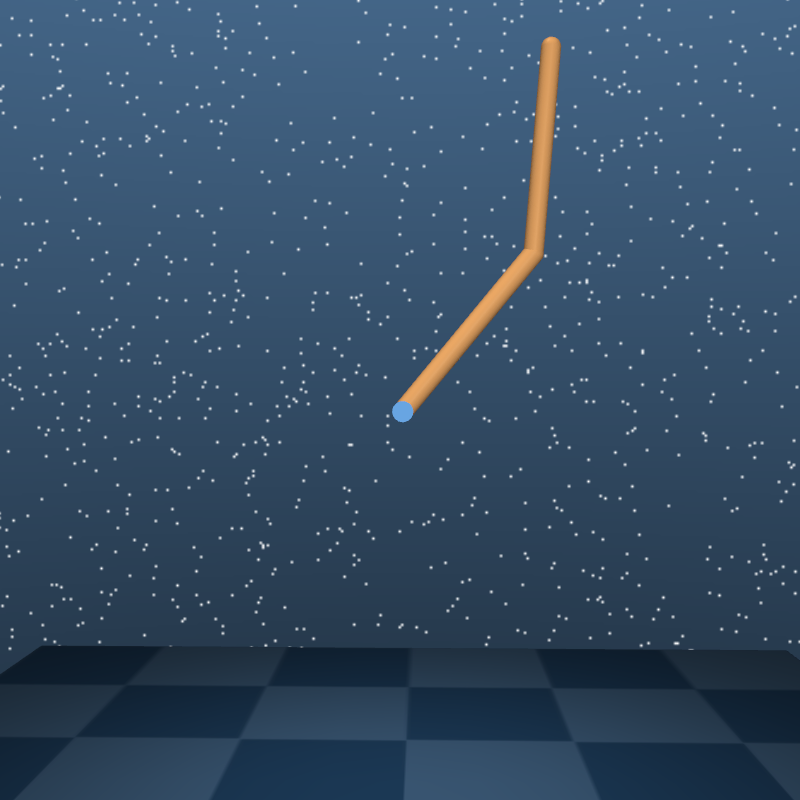}
    \includegraphics[width=0.184\textwidth]{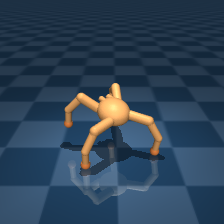}
    \includegraphics[width=0.184\textwidth]{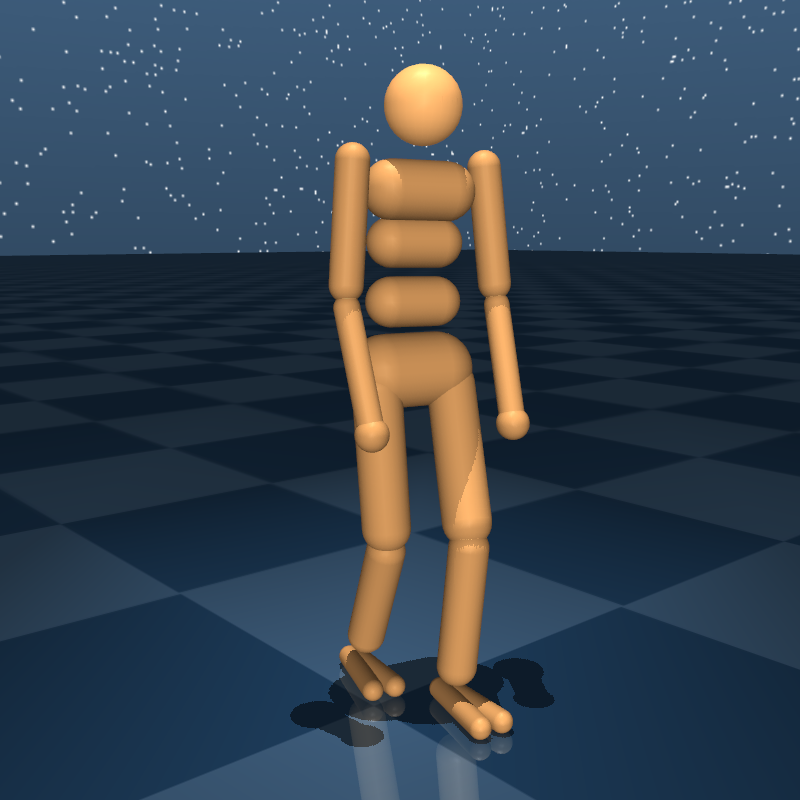}
    \caption{\textbf{Tasks visualizations.} We select some representative tasks from the four considered benchmarks. \textit{Top}: pen-v1, door-v1, hammer-v1, basketball-v2, hammer-v2. \textit{Bottom}: maze2d-large, arcobot-swingup, quadruped-walk, humanoid-stand.}
    \label{fig:task_visualizations}
    \vspace{-0.20in}
\end{figure}

\begin{table}[h!]
\caption{\textbf{Task configuration and hyperparameters.} We allocate tasks from three considered benchmarks into three task sets. Action repeat of 4 is widely applicable except for tasks with complex dynamics or shorter episode length. Following iCEM, we unify the colored-noise exponent to be 2.5 for smoother action sequences and 0.5 for high-frequency control tasks. The $\lambda$ coefficient varies across tasks for optimal performance. However, there are hints for this parameter tuning. For dense reward tasks where the estimation variance is dominant, a low coefficient is preferred. Conversely, a higher coefficient is better for the sparse reward tasks to accelerate the credit assignment process.}
\label{Table:per_task_settings}
\centering
\resizebox{0.8\textwidth}{!}{%
\begin{tabular}{llcccc} \toprule
    Environment set & Domain & Task & Action repeat & colored-noise & TD-${\lambda}$ \\ \midrule
    \textbf{Dense reward} & acrobot & swingup & ${4}$ & 2.5 & 0.8 \\ 
     & cheetah & run & ${4}$ & 0.5 & 0.4 \\ 
     & finger & turn\_hard & ${4}$ & 2.5 & 0.2 \\
     & fish & swim & ${4}$ & 2.5 & 0.2 \\ 
     & hopper & hop, stand & ${4}$ & 2.5 & 0.4 \\ 
     & humanoid & run, walk, stand & ${2}$ & 2.5 & 0.4 \\
     & pendulum & swingup & ${4}$ & 0.5 & 0.2 \\ 
     & quadruped & walk, run & ${4}$ & 0.5 & 0.2 \\ 
     & reacher & hard & ${4}$ & 2.5 & 0.2 \\ 
     & swimmer & swimmer6 & ${4}$ & 0.5 & 0.2 \\
     & walker & run & ${2}$ & 2.5 & 0.4 \\ \midrule
     \textbf{Primary} & acrobot & swingup-sparse & $4$ & 2.5 & 0.8 \\
     \textbf{goal-reaching} & adroit & pen & 2 & 0.5 & 0.2 \\ \midrule
     \textbf{long-horizon} & adroit & hammer, door & 2 & 2.5 & 0.2 \\
     & meta-world & all considered & 2 & 2.5 & 0.8 \\ \bottomrule
\end{tabular}%
}
\end{table}

\section{Extended Environmental Details} \label{app:envs}
We evaluate methods across four benchmarks, including D4RL~\citep{d4rl/20}, DMControl~\citep{dmc/18}, Adroit~\citep{dapg/18} and Meta-World~\citep{metaworld/19}. In this section, we make a detailed description of the environmental setup for the selected tasks. \cref{fig:task_visualizations} visualizes sampled frames from some of these tasks.

\subsection{D4RL}\label{app:d4rl_dateset}
We selected the maze2d-large environment to visually assess the agent's exploration ability. In this setting, the agent maneuvers a point mass using a 2-dimensional velocity command to navigate a maze without external rewards. Exploration ability is quantified using the maze coverage metric. We set the episode length to ${600}$ with an action-repeat of ${2}$. Additionally, we utilize the publicly available offline dataset from \url{http://rail.eecs.berkeley.edu/datasets/offline_rl/maze2d/maze2d-large-sparse-v1.hdf5} to test model prediction accuracy. This dataset, comprising ${4}$ million environment transitions, is pre-collected by a hand-designed navigation planner.

\subsection{DMControl}\label{app:dmc15_full}
\begin{figure*}[h!]
\centering
\includegraphics[width=0.95\textwidth]{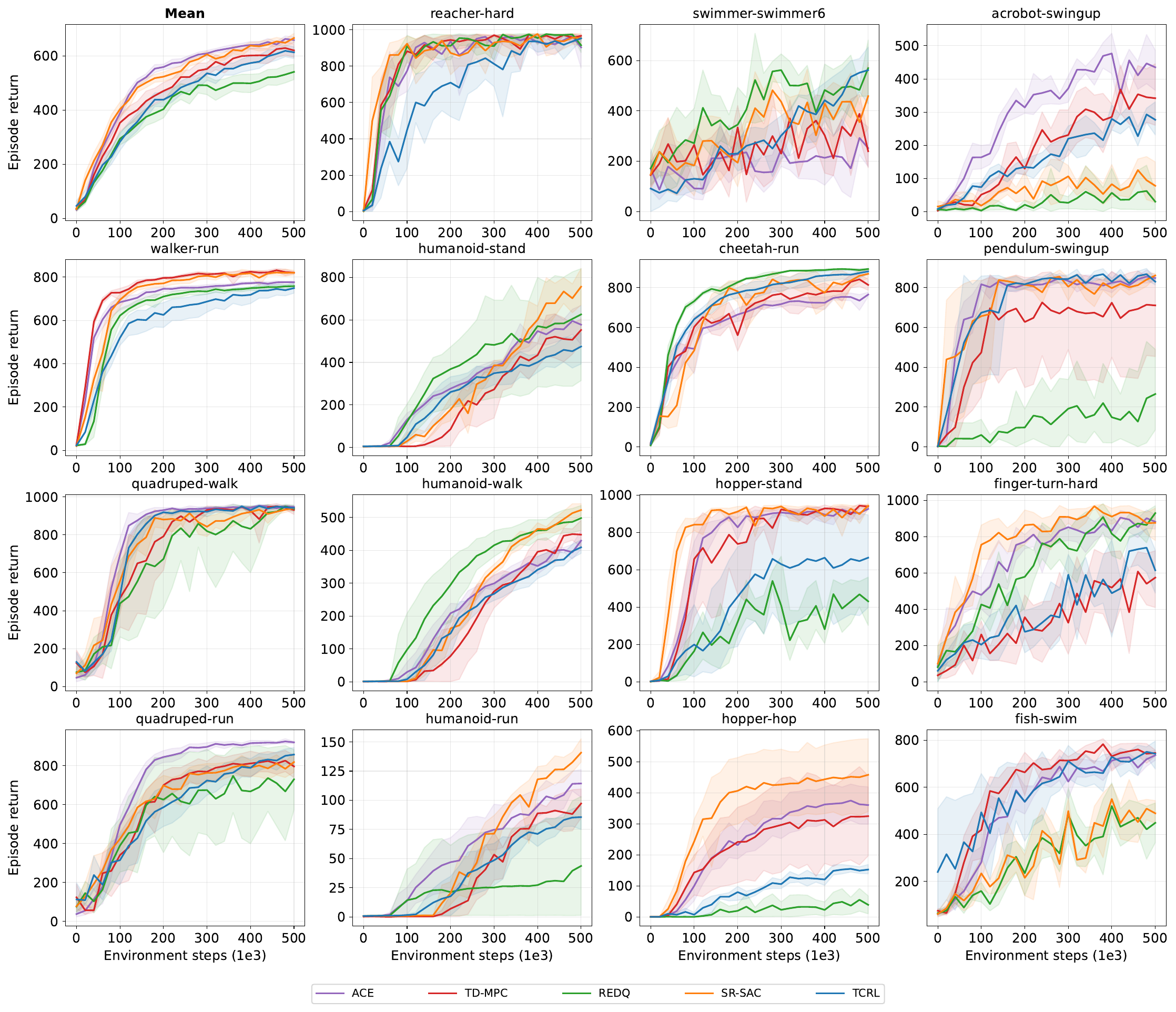} \vspace{-0.1in}
\caption{\textbf{DMC15-500k full results.} Episode return comparison between the proposed ACE planner and four sample-efficient baselines across 15 DMControl tasks.
Environment steps are constrained to be $500 \times 10^3$ for all tasks. Mean of 5 runs; shaded areas are 95\% CIs. Mean aggregated results, plotted in the top left, across tasks showcase the consistent overall performance of the ACE planner.}
\label{fig:dmc15_full_results}
\vspace{-0.15in}
\end{figure*}

We assess our method on 15 state-based tasks from DMControl, comprising 13 dense reward tasks and 2 primary goal-reaching ones. Task selection is based on the difficulty level, where the optimal solution is not immediately achievable by the baseline SAC, as indicated by the benchmark results at \url{https://github.com/denisyarats/pytorch_sac#results}.

For dense reward cases, we use the provided shaped rewards ${r(s, a) \in [0, 1]}$ that range in the unit interval. The episode length is set to 1,000 and consistent across all environments. Sparse reward tasks provide a per-step reward of 1 upon reaching the goal state and 0 otherwise. We leverage the specialized DMC15-500k benchmark~\citep{sac-RR/23} for dense reward cases, enabling fair sample efficiency comparison with ${5 \times 10^5}$ environment steps. This constraint is occasionally relaxed in challenging tasks to report learning progress in the main paper. Although two sparse reward tasks are incorporated, we assume that this benchmark provides a rigorous evaluation of dense reward performance because the primary goal-reaching can be relatively easier to solve. \cref{Table:per_task_settings} lists the task selection and their action-repeat configuration.

\textbf{Full results}\ \ We present the DMC15-500k full results in~\cref{fig:dmc15_full_results}, which includes a mean-aggregated result across all tasks. We find that the sample-efficient baselines generally perform well except for tasks with complex dynamics (i.e., \textit{humanoid-run} and \textit{acrobot-swingup}) or exploration bottlenecks (i.e., \textit{hopper-hop} and \textit{finger-turn-hard}). Obviously, the agents derive large performance gains from the lookahead search ability in these complex dynamics tasks. As a result, the ACE planner and TD-MPC achieve nearly four times the episode return compared to REDQ and SR-SAC, which use a policy for inference. In scenarios where shaped dense rewards are insufficient for finding a global optimal policy, employing a high UTD ratio along with model parameters reset is an effective strategy to escape local optima. Instead, the ACE planner employs the forward predictive intrinsic reward and MVE-based value estimation to get comparable performance with less computational intensity. Finally, we observe that proper state representation is always beneficial to the agent's final performance.

\subsection{Adroit} \label{subsec:adroit_env}
\begin{figure*}[h!]
\centering
\includegraphics[width=0.85\textwidth]{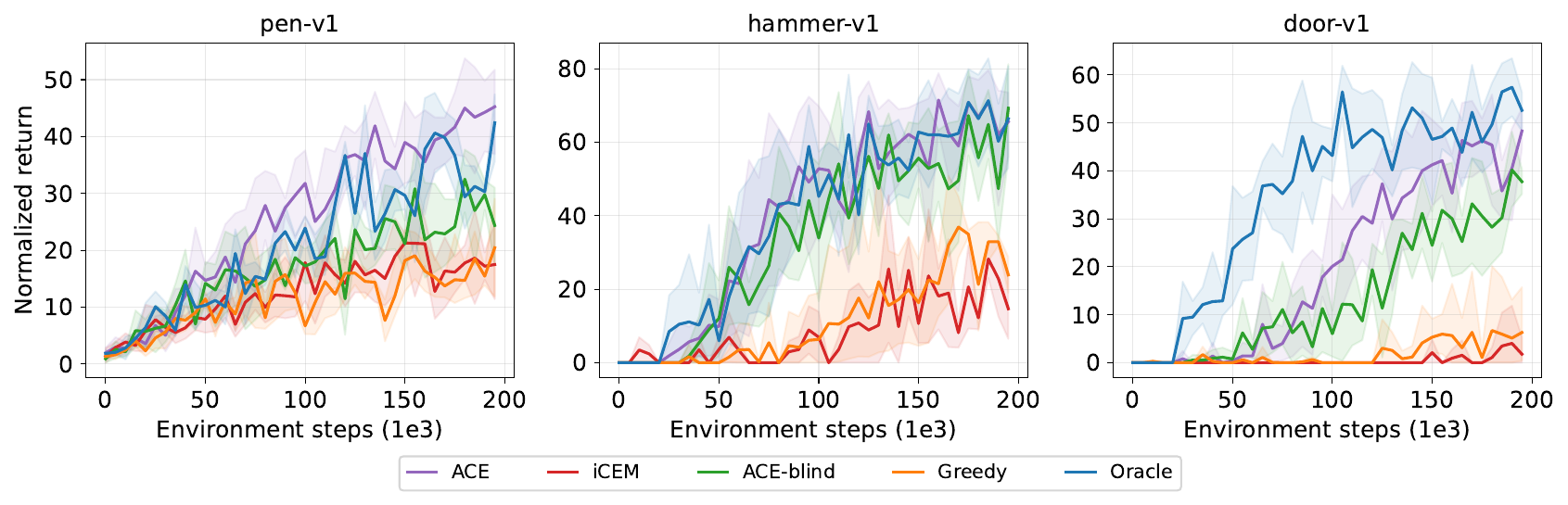} \vspace{-0.1in}
\caption{\textbf{Adroit sparse full results.} Normalized goal-reaching counts of the ACE planner and the other four exploration baselines on three Adroit sparse reward tasks. Mean of 5 runs; shaded areas are 95\% CIs. Notably, the ACE planner demonstrates comparable performance to the Oracle agent, which has access to shaped dense rewards.}
\label{fig:adorit_full_results}
\vspace{-0.15in}
\end{figure*}

We consider three sparse reward tasks from Adroit: Pen-v1, Door-v1, and Hammer-v1. Adhering to the benchmark's fundamental setup, we set episode lengths at 100 for Pen and 200 for Door and Hammer, with a consistent action repeat of 2 for all three tasks. We allocate the pen spin task to the primary goal-reaching set because the pen's goal state can be reached without hand movement, but just finger manipulation skills. Conversely, door and hammer tasks are characterized as long-horizon decision-making tasks, demanding proper hand movements to reach sequential sub-goals.

Sparse rewards for the primary goal-reaching set provide a per-step reward of 1 upon goal-reaching and 0 otherwise. In contrast, for the long-horizon tasks, we assign sparse rewards once each sub-goal is reached, leveraging a HER buffer to facilitate reaching the final goal. Specifically, we implement the HER buffer using the future goal selection strategy with a replay random states of 4. As for the oracle agent, the shaped reward functions are provided for each task. Further details on reward design are available in our open-source codebase at~\url{https://anonymous.4open.science/r/ace-torch-7868}.

For evaluation, we adopt a rigorous metric aligned with~\cite{rlpd/23}, where the performance score is characterized by the reach count of the final goal rather than the success rate. To report aggregated results across tasks within the benchmark, we normalize the scores to be the percentage of goal-reaching counts in the total timesteps. Consequently, this metric effectively reflects the task completion speed, providing a more discriminative evaluation. Finally, we find that a total of ${2 \times 10^5}$ environment steps and an action repeat of 2 are appropriate for distinguishing the performance gaps.

\textbf{Full results}\ \ The learning progress for individual Adroit tasks is shown in~\cref{fig:adorit_full_results}. Overall, the ACE planner performs comparably to the Oracle agent across tasks and notably outperforms in~\textit{pen-v1}. We attribute this superior performance to a mismatch between the shaped reward and desired agent behavior. Indeed, the dexterous manipulation skills required in the pen spin task are challenging to capture in a hand-shaped reward function. Instead, leveraging a sparse goal-reaching reward combined with the predictive intrinsic reward mitigates this mismatch, leading to enhanced performance. For \textit{hammer-v1} and \textit{door-v1}, intrinsic reward contributes modestly, as these tasks heavily rely on long-horizon decision-making ability, which is solved using the HER buffer. To summarize, the novelty-aware value function and H-step lookahead search ability are crucial to sample-efficient exploration, while the introduced intrinsic reward focuses on local dynamics transitions, accounting for model uncertainty.

\subsection{Meta-World}\label{app:meta-world_full_results}
\begin{figure*}[h!]
\centering
\includegraphics[width=0.85\textwidth]{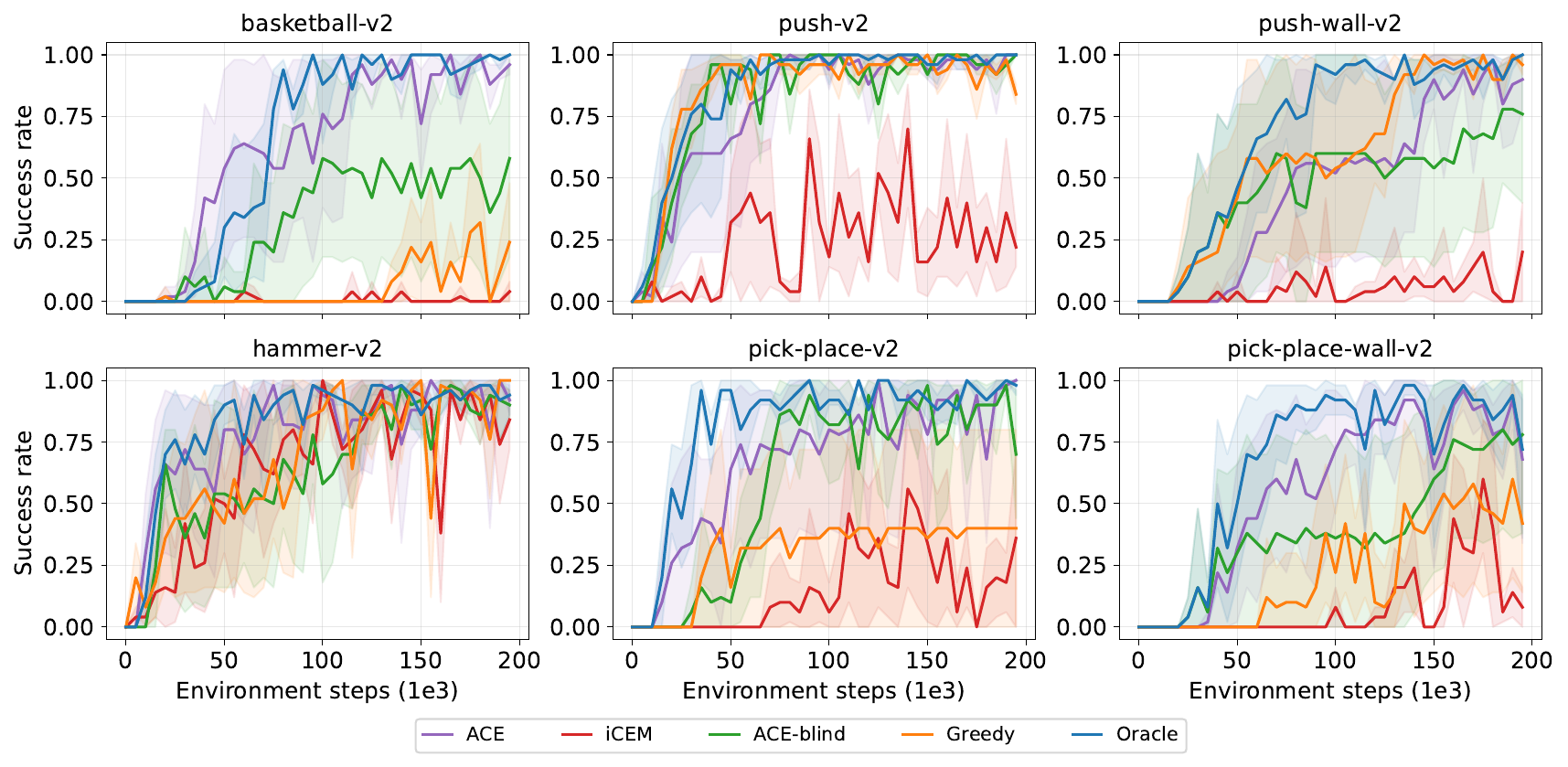} 
\caption{\textbf{Meta-World sparse full results.} Success rate of the ACE planner and the other four exploration baselines on $6$ Meta-World sparse reward tasks. Mean of $5$ runs; shaded areas are $95\%$ CIs.}
\label{fig:meta-world_full_results}
\end{figure*}

We specifically choose $6$ sparse reward tasks from Meta-World, with an emphasis on tasks that involve long-horizon decision-making. For instance, the object manipulation ability followed by a proper gripper movement is required in \textit{pick-place} task. To focus on the efficient exploration property of the evaluated agents, we configure Meta-World environments to be tasks with visible single goals. Consistently, we set the action repeat to $2$ and the episode length to $100$ across all tasks, which is considered sufficient to reach all final goals.

Sparse reward design mirrors that of Adroit, and a HER buffer is also incorporated. To facilitate a strict comparison in the low-data regime, we constraint the total environment steps to be $2 \times 10^5$, in which the Oracle agent is capable of solving all selected tasks. We adapt the success rate metric in which the agent is deemed to be successful if it finishes the episode closing to a goal state.

\textbf{Full results}\ \ The learning progress and success rates for each selected Meta-World task are presented in~\cref{fig:meta-world_full_results}. We generally observe that these single-goal tasks can be solved by the Oracle agent, and the ACE planner is competitive under sparse reward conditions. This suggests that, in these specific tasks, sophisticated reward shaping can be replaced with hindsight goal-reaching without compromising sample efficiency. In scenarios involving complex manipulation, such as \textit{basketball-v2, push-wall-v2, pick-place-wall-v2}, the ACE-blind baseline exhibits an inferior success rate compared to the ACE planner due to more frequent failure cases.

\section{Extended Ablation Results}
\textbf{Relative importance of each component.} Here, we provide the full ablation results involving useful combinations of critical design choices, including an improved CEM planner, intrinsic reward, and the MVE-based value estimator. Evaluation metrics encompass episode return, the average normalized estimation error of the value function, and its standard deviation, which provide a detailed insight into the quality of the value estimation quality~\citep{REDQ/21}.

\begin{figure*}[h!]
\centering
\includegraphics[width=0.85\textwidth]{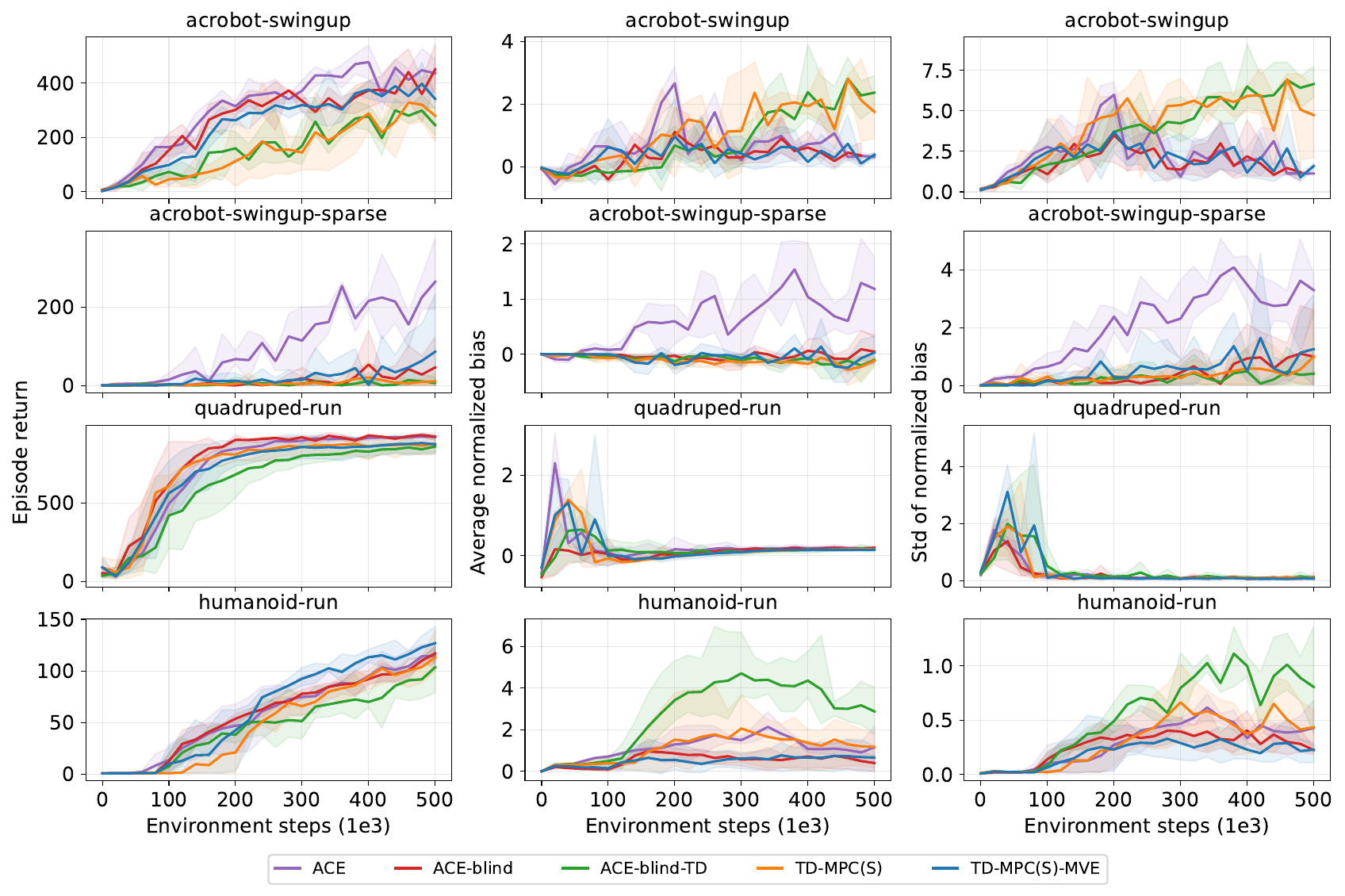} \vspace{-0.10in}
\caption{\textbf{Ablation results on critical design choices.} Evaluation of ACE, ACE-blind, and TD-MPC(S)-MVE, each utilizing the MVE-based value target. TD-MPC(S)-MVE and TD-MPC(S) opt for a vanilla CEM planner in contrast to others employing the iCEM planner. The intrinsic reward design is exclusive to our method, ACE. Mean of 5 runs; shaded areas represent 95\% CIs.}
\label{fig:estimate_bias_full}
\vspace{-0.15in}
\end{figure*}

As depicted in~\cref{fig:estimate_bias_full}, agents employing the MVE-based value target generally show superior performance across the considered tasks with lower averaged estimation bias and its std of the value function. Besides, the MVE-based value estimator is planner-agnostic and sometimes gets a better result when combined with the CEM planner. This indicates that the performance enhancement in our method is not derived from the replacement of the online planner but is significantly influenced by the modified value target. Of particular note, only the ACE planner demonstrates non-trivial performance on \textit{acrobot-swingup-sparse}, attributed to the incorporation of the proposed intrinsic reward. Importantly, the intrinsic reward proves versatile across the considered tasks, contributing to improved sample efficiency and asymptotic performance.

\label{app:extended_ablation_results}

\textbf{Hypperparmeters.} \ \
\begin{figure*}[h!]
\centering
\includegraphics[width=0.85\textwidth]{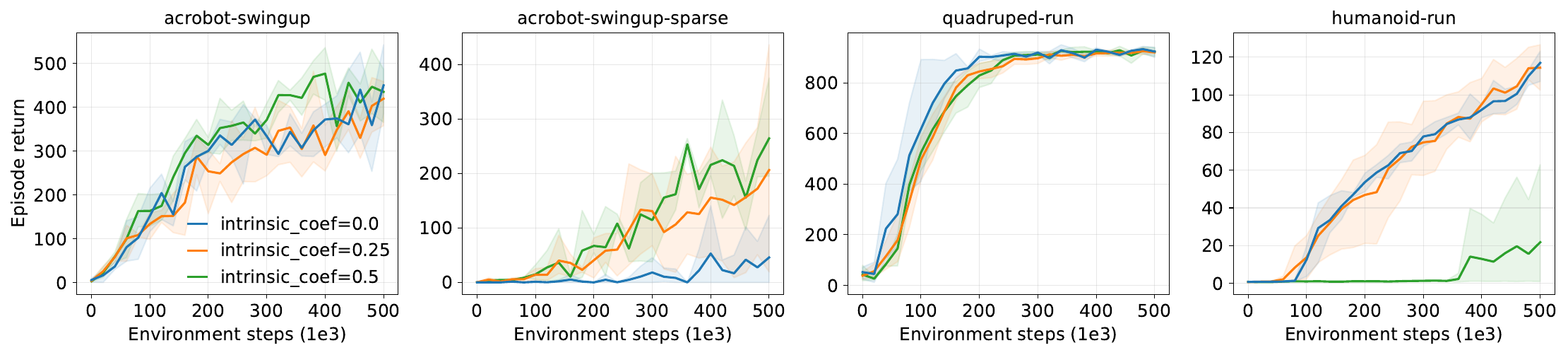} \\ \vspace{0.05in}
\includegraphics[width=0.85\textwidth]{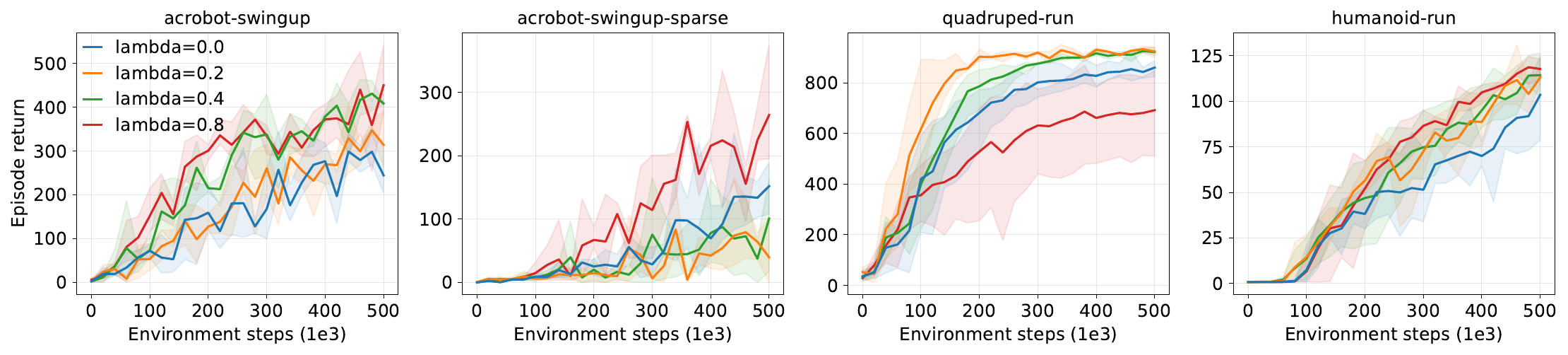} 
\vspace{-0.10in}
\caption{\textbf{Core hyperparameters ablation.} \textit{(Top)} Influence of varying the intrinsic coefficient in the range $[0.0, 0.25, 0.5]$. \textit{(Bottom)} Investigation of the lambda coefficient for the ACE planner, illustrating the tradeoff between estimation bias and std of the value function. Mean of $5$ runs; shaded areas represent 95\% CIs.}
\label{fig:hyper-parameters_ablation}
\vspace{-0.15in}
\end{figure*}

In this section, we conduct a grid search of two core hyperparameters, namely the intrinsic reward and lambda coefficients, to uncover the parametric sensitivity of the ACE planner. We first select the intrinsic reward coefficient from $[0.0, 0.25, 0.5]$, which is abbreviated as \textit{intrinsic\_coef} in the top row of~\cref{fig:hyper-parameters_ablation}. The ACE planner exhibits parameter insensitivity in dense reward tasks with modest difficulty levels, such as \textit{acrobot-swingup} and \textit{quadruped-run}. However, the impact of this coefficient is more pronounced in sparse reward tasks, where a larger value is favored, while a smaller one is preferred for challenging high-dimensional control tasks. Consequently, we recommend a unified intrinsic reward coefficient of 0.25 for dense reward tasks and 0.5 for sparse ones.

The lambda is a crucial coefficient that controls the weights assigned to the estimated value targets with different model rollout horizons, establishing a tradeoff between estimation bias and std of the value function. When setting lambda to be $0$, the MVE-based value target reduces to a one-step TD target. As shown in the bottom row of~\cref{fig:hyper-parameters_ablation}, we vary the coefficient value from $[0.0, 0.2, 0.4, 0.8]$ and find that there is no silver bullet across the considered tasks. A modest weight assignment, specifically setting lambda to be $0.2$ or $0.4$, often achieves comparable or better performance to the vanilla agent. However, setting the lambda to $0.8$ may attain significant performance improvement in tasks requiring long-term credit assignment, such as the \textit{acrobot-swingup-sparse} task.

\end{document}